\pdfoutput=1
\documentclass[11pt]{article}
\usepackage[]{acl}
\usepackage{listings}

\usepackage{times}
\usepackage{placeins}
\usepackage{latexsym}
\usepackage{lipsum}
\usepackage{booktabs}
\usepackage[labelfont=bf]{caption}
\usepackage{multirow}
\usepackage{graphicx}
\usepackage{amssymb}
\usepackage{amsmath}
\usepackage{enumerate}
\usepackage{soul}
\usepackage{booktabs}   
\usepackage{xcolor}     
\usepackage{array}      
\usepackage{enumitem}
\usepackage[T1]{fontenc}

\usepackage[utf8]{inputenc}

\usepackage{microtype}

\usepackage{inconsolata}

\usepackage{graphicx} 
\usepackage{natbib}  
\usepackage{caption} 
\usepackage{multirow}
\usepackage{algorithm}
\usepackage{algpseudocode}
\usepackage{amsmath}
\usepackage{booktabs}
\usepackage{microtype}

\usepackage{amsfonts}
\usepackage{blkarray, bigstrut}
\parskip=\medskipamount
\usepackage{amssymb}
\usepackage{pifont}
\newcommand{\gemmaone}{{\tt Gemma-1.1-7b-it}}

\newcommand{\mistralinstructv}[2]{{\tt Mistral-$#1$B-Instruct-v$#2$}}

\newcommand{\gemmatwo}{{\tt Gemma-2-9b-it}}

\newcommand{\qweninstructnew}{{\tt Qwen/Qwen2-7B-Instruct}}

\newcommand{\mixtralnew}{{\tt mistralai/Mixtral-8x22B-Instruct-v0.1}}

\usepackage[labelfont=bf]{caption}
\usepackage{multirow}

\usepackage{newfloat}
\usepackage{listings}

\usepackage{xcolor}
\usepackage[capitalise,noabbrev]{cleveref}
\usepackage{hyperref}
\usepackage{xcolor}
\usepackage{multicol}

\definecolor{myYellow}{RGB}{255,255,0}
\definecolor{myBlue}{RGB}{0,0,255}
\definecolor{myGreen}{RGB}{0,128,0}
\definecolor{myRed}{HTML}{D62728}       
\definecolor{myGreen}{HTML}{2CA02C}     
\definecolor{myBlue}{HTML}{1F77B4}      
\definecolor{myOrange}{HTML}{FF7F0E}    

\newcommand\blfootnote[1]{%
  \begingroup
  \renewcommand\thefootnote{}\footnote{#1}%
  \addtocounter{footnote}{-1}%
  \endgroup
}

\title{Why We Feel What We Feel: Joint Detection of Emotions and Their Opinion Triggers in E-commerce}

\author{Arnav Attri$^\diamondsuit$$^\clubsuit$,
Anuj Attri$^\diamondsuit$$^\clubsuit$, Pushpak Bhattacharyya$^\clubsuit$ 
\\ 
\textbf{Suman Banerjee$^\mathcal{F}$, Amey Patil$^\mathcal{F}$, Muthusamy Chelliah$^\mathcal{F}$, Nikesh Garera$^\mathcal{F}$}\\
\textbf{}
        $^\clubsuit$Computer Science and Engineering, IIT Bombay, India, 
        $^\mathcal{F}$Flipkart, India \\
        \texttt{\{arnavcs, ianuj,
        pb\}@cse.iitb.ac.in}
        }

\begin{document}
\maketitle
\blfootnote{$^\diamondsuit$ Equal contribution}
\begin{abstract}
Customer reviews on e-commerce platforms capture critical affective signals that drive purchasing decisions. However, \textit{no} existing research has explored the joint task of emotion detection and explanatory span identification in e-commerce reviews - a crucial gap in understanding what triggers customer emotional responses. To bridge this gap, we propose a novel joint task unifying \textsc{\textbf{E}motion} detection and \textsc{\textbf{O}pinion \textbf{T}rigger} extraction (\textsc{\textbf{EOT}}), which explicitly models the relationship between causal text spans (opinion triggers) and affective dimensions (emotion categories) grounded in Plutchik's theory of $8$ primary emotions.
In the absence of labeled data, we introduce \textsc{\textbf{EOT-X}}, a human-annotated collection of \textbf{2,400} reviews with fine-grained emotions and opinion triggers. We evaluate $\mathbf{23}$ Large Language Models (LLMs) and present \textsc{\textbf{EOT-DETECT}}, a structured prompting framework with systematic reasoning and self-reflection. Our framework surpasses zero-shot and chain-of-thought techniques, across e-commerce domains.


\end{abstract}

\section{Introduction}

Emotional content in customer reviews fundamentally drives consumer behavior \cite{chen2022impact} demonstrates their direct influence on purchasing decisions, while \cite{baba} establishes their impact on post-purchase satisfaction. These emotional expressions powerfully shape how future consumers perceive and engage with products \cite{greifeneder2007extending, pham2007emotion, kim2019emotion, wang2022, bian2022analyzing}.

Despite these clear benefits, \textit{no} existing research in literature has explored the joint task of emotion detection and explanatory span identification in e-commerce reviews, leaving a significant gap in understanding \textit{what} customers feel and \textit{why} they feel it.


For instance, merely detecting "\textit{disgust}" in a review provides limited value compared to identifying that this emotion stems from specific triggers explicitly stated by customers, such as \textit{finding hair strands in makeup products or discovering expired skincare items.}

We address this gap by introducing Emotion detection and Opinion Trigger extraction (EOT), a joint task that directly models the bidirectional relationship between emotions and their corresponding opinion triggers, grounded in Plutchik's theory \cite{plutchik1980emotion}.

\begin{figure}[t]
    \centering
    \includegraphics[width=1\columnwidth]{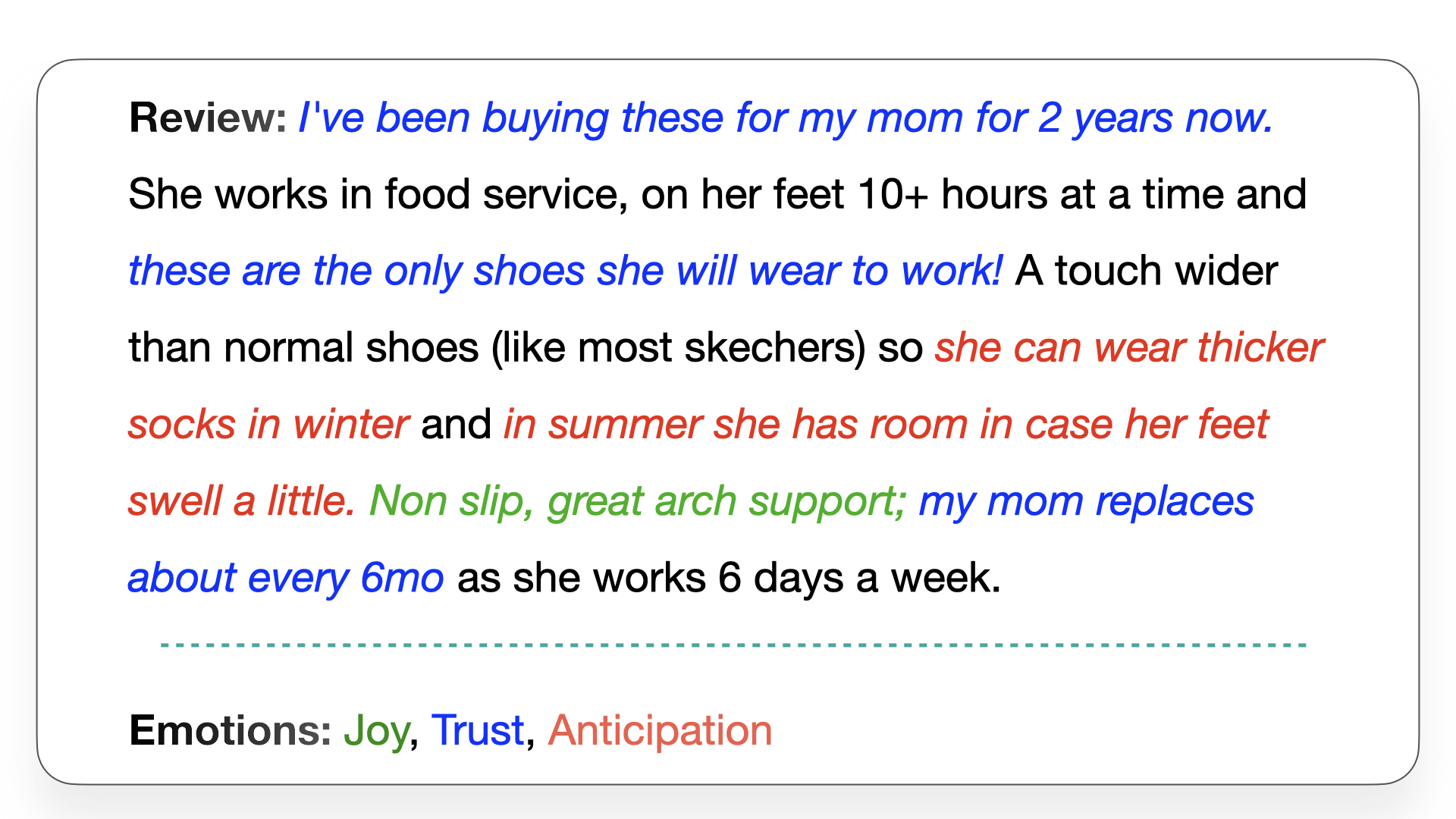}
    \caption{EOT-X dataset example: Product review with emotion triggers highlighted by color to indicate corresponding emotions. Each trigger is a contiguous span showing the emotion's cause.}
\label{fig:first}
\end{figure}

\paragraph{Why Plutchik's $8$ Primary Emotions}
Several factors guided our choice of Plutchik's 8 primary emotions:


\noindent(\textit{1}) {\tt \textbf{LLM Behavior Control}}: In our preliminary experiments, unconstrained LLMs generated redundant emotion labels (synonyms and variants) of basic emotions, creating an \textit{"emotion explosion}" that hindered systematic evaluation and comparison. Plutchik's framework enforces taxonomic constraints without compromising expressivity (Appendix X).

\noindent(\textit{2}) {\tt \textbf{Theoretical Strength}}: Plutchik's model balances positive and negative emotions, backed by psychological and empirical research. Prior work \cite{strapparava-mihalcea-2007-semeval} often limited analysis to 6 emotions or focused mainly on negative ones.

\noindent(\textit{3}) {\tt \textbf{Balanced Scope}}: \cite{ekman1992argument} identifies $6$ basic emotions, but Plutchik's model adds granularity by including {\tt trust} and {\tt anticipation}, which are crucial for analyzing customer feedback.

\noindent(\textit{4}) {\tt \textbf{Practical Implementation}}: Consistent with \cite{mohammad2013crowdsourcing}, our observations show that annotating hundreds of emotions is costly and cognitively demanding.



\textls[-60]{\textsc{\textbf{Opinion Trigger}}} is a \textit{verbatim text} segment from a review that directly shows what caused a customer's emotional response, allowing verification of the connection between emotions and their specific causes.


Our contributions are:
\begin{enumerate}
    \item \textsc{\textbf{Emotion-Opinion Trigger (EOT):}} The pioneering joint task that leverages LLMs to unify fine-grained emotion detection and opinion trigger extraction from \textit{customer reviews}, enabling interpretable and actionable analysis of user feedback (Section \ref{task_formulation}).

    \item \textsc{\textbf{EOT-X:}} The first human-annotated benchmark dataset (Section \ref{annotation_dataset}) of $\mathbf{2,400}$ reviews curated from Amazon, Yelp, and TripAdvisor to support cross-domain generalization. Each review is labeled with fine-grained emotions from Plutchik’s $8$ primary emotions and extractive opinion triggers identifying the emotion sources, enabling model fine-tuning.
    

    \item \textbf{EOT-LLAMA:} An edge-deployable LLM (fine-tuned from {\tt LLAMA-3.2.1B-INSTRUCT}) designed to detect Plutchik's $8$ emotions and extract opinion triggers from customer reviews. No existing model addresses this dual task. It outperforms models up to $7$x larger counterparts (Table \ref{tab:ft_models}) while remaining deployable on \textit{consumer-tier hardware}.

    \item \textbf{EOT-DETECT:} A structured prompting framework that leverages systematic reasoning and self-reflection steps (Section \ref{ok}, \ref{result}), achieving superior performance over zero-shot and chain-of-thought.

    \item \textsc{\textbf{Comprehensive Benchmarking:}}  We conduct the first large-scale evaluation of $23$ LLMs (closed- and open-source) for joint emotion detection and opinion trigger extraction on customer reviews (Table \ref{tab:results}, \ref{tab:closed_results}).

\end{enumerate}


\textit{To the best of our knowledge}, we present the first unified framework for joint emotion-opinion analysis, conducting model comparisons at an unprecedented scale. We will publicly release our curated benchmark dataset, and fine-tuned model as \textit{foundational contributions} to the research community.

\section{Related Work}

\textls[-60]{\textsc{\textbf{Emotion Analysis}}} is crucial in natural language processing, demonstrating how emotions shape online discourse, decision-making, and user behavior, particularly in e-commerce where customer reviews influence purchasing decisions \cite{mohammad2013crowdsourcing, malik2024nlp}. While early approaches focused on sentiment polarity \textit{(negative, neutral, positive)}, researchers recognized the need for nuanced emotion taxonomies to capture human emotional expression complexity \cite{russell1980circumplex, ekman1992argument, plutchik2001nature}.


\textls[-60]{\textsc{\textbf{Emotion-Trigger Analysis}}} remains an under-explored dimension of emotion understanding. Early research utilized rule-based approaches (\cite{neviarouskaya2009compositionality, lee-etal-2010-text} and statistical methods \cite{gui-etal-2016-event, xia-ding-2019-emotion}
 to identify emotion causes in text. Recent studies examined graph-based models and attention mechanisms for joint emotion-cause extraction \cite{wei-etal-2020-unified, fan-etal-2021-joint, singh2021end} and context-aware trigger identification \cite{li2019context}. Studies demonstrate the complexity of linking emotional expressions to triggers. \cite{singh-etal-2024-language}
 presented EMOTRIGGER dataset, testing LLMs on social media data, showing their limitations in trigger identification despite strong emotion recognition. \textit{Emotion-trigger analysis remains unexplored in e-commerce.}

\textls[-60]{\textsc{\textbf{Dataset Development}}} has driven emotion analysis research advancement. Key datasets include SemEval \cite{strapparava-mihalcea-2007-semeval}, GoEmotions \cite{demszky-etal-2020-goemotions}, and ISEAR \cite{scherer1994isear}. Domain-specific datasets like CancerEmo \cite{sosea-caragea-2020-canceremo} and EmoCause \cite{gui-etal-2016-event} emerged. \textit{However, \textbf{no }existing dataset provides emotion labels with trigger annotations for e-commerce platforms.}

\begin{figure*}[htp]
    \centering
    \includegraphics[width=2\columnwidth]{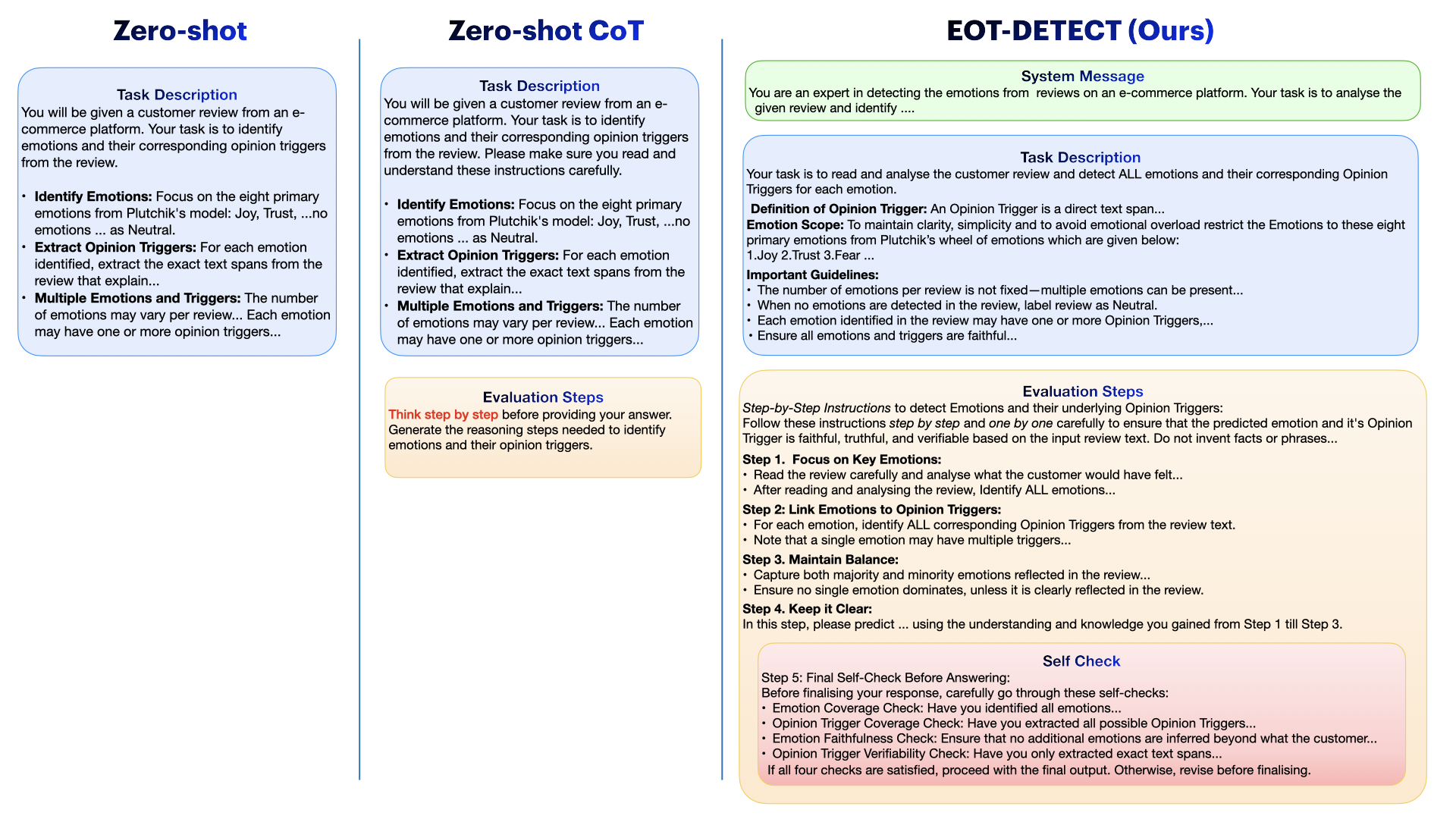}
    \caption{This figure illustrates three prompting strategies for joint emotion detection and opinion trigger extraction from e-commerce reviews: Zero-Shot (ZS), Zero-Shot Chain-of-Thought (ZS-CoT), and our proposed EOT-DETECT. ZS provides a basic task description. ZS-CoT adds a "think step-by-step" instruction to encourage reasoning. EOT-DETECT structures the prompt with a System Message (defining the LLM's role), a detailed Task Description (specifying constraints), and a 5-step Instruction sequence (guiding the reasoning process). Critically, EOT-DETECT incorporates a final Self-Check step to ensure comprehensive emotion coverage, accurate trigger extraction, and verifiable results.}
    \label{fig:big}
\end{figure*}
\textls[-60]{\textsc{\textbf{LLMs}}} have shown exceptional capabilities in understanding and generating emotionally nuanced text \cite{brown2020language, ouyang2022training}. Research has explored their potential for emotion analysis \cite{acheampong2023text, huang2024feeling}, demonstrating success in zero-shot and few-shot settings. \textit{The application of LLMs to combined emotion detection and trigger identification in e-commerce remains unexplored.} 

While prior research has significantly advanced emotion analysis and trigger detection, a critical gap remains in analyzing customer feedback within e-commerce contexts—\textit{a gap that our work seeks to bridge.}

\section{Task Formulation}\label{task_formulation}
We define the \textit{key} components as:

\subsection{Review and Emotion Representation}

Let $\mathcal{R} = \{R_i\}_{i=1}^{N}$ denote a review consisting of $\mathbf{N}$ tokens. We focus on the set of eight primary emotions from Plutchik's model: $\mathcal{E} = \{\tt{Joy},$ $\tt{Trust},$ $\tt{Fear},$ $\tt{Surprise},$ $\tt{Sadness},$ $\tt{Disgust},$ $\tt{Anger},$ $\tt{Anticipation}\}$, with an additional {\tt{Neutral}} label for cases in which no discernible emotion is expressed. Thus, our extended set of possible emotions is: $\widehat{\mathcal{E}} = \mathcal{E} \cup \{\tt{Neutral}\}$.

\subsection{Emotion Detection and Trigger Extraction}
A review can have multiple emotions and a single emotion can have many opinion triggers.
We \textit{seek} to detect emotions in $\widehat{\mathcal{E}}$ present in $\mathcal{R}$ and extract their corresponding  \textls[-60]{\textsc{Opinion Triggers}} characterized as $ \mathcal{T}_e = \{\, (i, j)\,\mid\, (w_i, \dots, w_j) \text{ is a substring explaining emotion } e\}\,$. If no emotions in $\mathcal{E}$ are detected, then we assign:
$
\mathcal{O}(\mathcal{R}) = \{(\tt{Neutral}, \varnothing)\}\,.
$
$\mathcal{T}_e$ extractive design enables direct verification against source reviews.

\subsection{Extractive Constraints}

Let $\mathcal{M}$ be an LLM that, given $\mathcal{R}$ and a structured prompt $\mathcal{P}$, solves:

\textls[-60]{\textsc{\textbf{1. Emotion Identification}}}: A subset $\mathcal{E}'$ of the $8$ primary emotions $\mathcal{E}$ expressed in $\mathcal{R}$ is selected by $ f_{\mathrm{emo}}(\mathcal{R}) \to \mathcal{E}' \subseteq \mathcal{E}$. If $\mathcal{E}' = \varnothing$, we label the review as {\tt{Neutral}}.

\textls[-60]{\textsc{\textbf{2. Opinion Trigger Extraction}}}: For each identified emotion $e_j \in \mathcal{E}'$, the model extracts a set of triggers: $f_{\mathrm{trig}}(\mathcal{R}, e_j) \to \mathcal{T}_j \subseteq \mathcal{S}(\mathcal{R}),$ where $\mathcal{S}(\mathcal{R})$ is the set of all possible contiguous substrings in $\mathcal{R}$. For evaluation and verification, each candidate trigger $t \in \mathcal{T}_j$ must satisfy the extractive constraint: $\forall t \in \mathcal{T}_j,\ \text{span}(t) \subseteq \mathcal{R}.$

\subsection{Output Representation}

The final outcome: $\mathcal{O}(\mathcal{R}) = \{(e,\mathcal{T}_e) \mid e \in \widehat{\mathcal{E}}, \mathcal{T}_e \neq \varnothing\}$ if at least one emotion from $\mathcal{E}$ is present, and $\{(\tt{Neutral},\varnothing)\}$ otherwise.

\section{Methodology}\label{method} 

Our methodology involves three approaches as shown in Figure \ref{fig:big}:

\subsection{Zero-Shot Prompting (ZS)}

In zero-shot prompting \cite{brown2020language}, we instruct the LLM to identify emotions and extract opinion triggers from a given review without any task-specific examples.

\subsection{Zero-Shot Chain-of-Thought (ZS-CoT)}

Following \cite{kojima2022large}, this approach explicitly prompts the LLM to \texttt{<think step-by-step>} before generating its final output, encouraging structured reasoning. A key distinction is that the \texttt{<reasoning>} block is \textit{generated} by the model, not provided as input, ensuring the model articulates its analysis before identifying emotion-trigger pairs.

\subsection{EOT-DETECT Framework} \label{ok}
We introduce \textsc{\textbf{EOT-DETECT}}, a structured prompting framework that guides LLMs to perform emotion-opinion trigger analysis on reviews with improved \textit{instruction compliance}. Inspired by insights from \textit{human} cognitive processes \cite{fiske1991social} , our framework interleaves \textit{human-like reasoning steps} and \textit{self-reflection mechanisms} {\tt (or “self-checks”)} to ensure extractive trigger identification and comprehensive emotion coverage.

Following our task formulation, we design a structured prompt $\mathcal{P}$ to enforce explicit constraints on both emotion detection and trigger extraction. Concretely, \textls[-30]{EOT-DETECT}
 is defined as a $4$-tuple:
$
\mathcal{P}= \langle \texttt{\textbf{S}}, \texttt{\textbf{T}}, \texttt{\textbf{I}}, \texttt{\textbf{R}} \rangle
$

\noindent(\textit{1}) {\tt \textbf{System Message (S)}}: Defines the model’s expert persona and sets the global context for how the model should interpret subsequent instructions.

\noindent(\textit{2}) {\tt \textbf{Task Description (T)}}: A structured component that delineates the scope and purpose of the prompt, composed of: $T = \langle \text{Task Definition}, \text{Opinion Trigger Definition}, \\
    \text{Emotion Scope}, \text{Guidelines} \rangle$

        \begin{itemize}
        \item \textbf{\tt Task Definition}: Explains that the model should detect emotions and extract Opinion Triggers directly from the text.
        \item \textbf{\tt Opinion Trigger Definition}: Clarifies that triggers must be exact substrings from \( R \).
        \item \textbf{\tt Emotion Scope}: Restricts focus to the $8$ Plutchik emotions plus {\tt{Neutral}}.
        \item \textbf{\tt Guidelines}: Reinforces constraints such as “no modifications” and “multiple emotions possible,” ensuring faithfulness to the review text.
    \end{itemize}

\noindent(\textit{3}) {\tt \textbf{Instructions (I)}}: A sequence of $\mathbf{5}$ carefully designed \textit{reasoning steps} that progressively guide the LLM through the analysis. 


Formally:  $I = \langle \texttt{I}_1, \texttt{I}_2, \texttt{I}_3, \texttt{I}_4, \texttt{I}_5 \rangle $ where,
    \begin{itemize}
        
        \item \textbf{\( \texttt{I}_1 \): {\tt Focus on Key Emotions}} – Directs the model to read the review carefully and identify all relevant emotions (or label as {\tt Neutral} if none are found).
        \item \textbf{\( \texttt{I}_2 \): {\tt Link Emotions to Opinion Triggers}} – For each emotion, prompts the LLM to extract one or more text spans from \( R \) explaining why the emotion is present.
        
        \item \textbf{\( \texttt{I}_3 \): {\tt Maintain Balance}} – Ensures that both majority and minority emotions are captured, so no single emotion is over- or underrepresented.
        \item \textbf{\( \texttt{I}_4 \): {\tt Keep it Clear}} – Instructs the model to finalize all detected emotions and triggers, adhering to \textbf{extractive} requirements.
        \item \textbf{\( \texttt{I}_5 \): {\tt Final Self-Check}} – A critical self-reflection step, prompting the LLM to verify \textbf{Emotion Coverage}, \textbf{Trigger Coverage}, \textbf{Emotion Faithfulness}, and \textbf{Opinion Trigger Verifiability} before producing the final output.
    \end{itemize}

\noindent(\textit{4}) {\tt \textbf{Input Review (R)}}: The text from which the model must \textbf{extract} triggers and identify emotions.

\paragraph{Prompt Execution}
When the structured prompt $\mathcal{P}$ is fed into an LLM (denoted by $\mathcal{M}$, the model sequentially processes the system message {\tt \textbf{(S)}}, the task description {\tt \textbf{(T)}}, the stepwise instructions {\tt \textbf{(I)}}, and finally the review {\tt \textbf{(R)}}. We denote this process as: $ M(\mathcal{P}) \longmapsto \mathcal{O}(\mathcal{R}) $, where \( \mathcal{O}(\mathcal{R}) \) is the emotion-trigger assignment defined in Task Formulation. 


\paragraph{Self-Reflection Mechanism}
A standout aspect of our framework is the \textls[-60]{\textbf{\( \texttt{I}_5 \) {\tt Final Self-Check}}}, which asks the model to confirm $4$ critical conditions before providing its final answer: (\textit{1}) \textls[-60]{\textbf{\texttt{Emotion Coverage Check}}}: All expressed emotions (including minority ones) must be captured.
(\textit{2}) \textls[-60]{\textbf{\texttt {Opinion Trigger Coverage Check}}}: All relevant triggers for each emotion are extracted, allowing multiple triggers for a single emotion.
(\textit{3}) \textls[-60]{\textbf{\texttt {Emotion Faithfulness Check}}}: No extra (unwarranted) emotions are added.
(\textit{4}) \textls[-60]{\textbf{\texttt {Opinion Trigger Verifiability Check}}}: Every trigger is exactly a sub-string of $\mathcal{R}$.

This step counters the tendency of LLMs to \textit{overgeneralize or hallucinate} content by explicitly requiring a final verification loop.

\section{Dataset}

We utilized multi-domain dataset \textls[-60]{$\mathcal{D}$} = $\{\tt{Amazon}, \tt{TripAdvisor}, \tt{Yelp}\}$, we employed \textls[-60]{Simple Random Sampling Without Replacement {\textsc(SRSWOR)}} \cite{cochran1977sampling}. The Amazon subset (from Amazon Reviews '23; \cite{hou2024bridging}
) spans subdomains $\mathcal{S} = \{\tt{Beauty}, \tt{Home}, \tt{Electronics}, \tt{Clothing}\}$. Let $P_s$ denote products in subdomain $s \in \mathcal{S}$. We sample $n_s = 40$ products per subdomain using \textsc{SRSWOR}: 
$
p_i \sim \textsc{SRSWOR}(P_s, n_s), \quad \forall s \in \mathcal{S},
$
yielding $\sum_{s \in \mathcal{S}} n_s = 160$ products. For TripAdvisor \cite{li-etal-2014-towards} and Yelp \cite{yelp_dataset}
, we sampled $n_t = 40$ items each using \textsc{SRSWOR}.  

For each product $p$, we define a length-filtered review set:$
\mathcal{R}_p = \{r \in \mathcal{R}'_p \mid 10 \leq \text{len}(r) \leq 100\}
$
to control for verbosity \citep{Kim2019-gs, Herrando2021-pm, Xie2022-rg}. We then sample $m = 10$ reviews using temporal stratification: $
\mathcal{R}_p^* \sim \textsc{SRSWOR}(\mathcal{R}_p, m) \quad \text{s.t.}\quad \forall t \in \mathcal{T}: |\mathcal{R}_p^* \cap \mathcal{R}_t| \propto |\mathcal{R}_t|.
$
This ensures: (\textit{1}) Uniform coverage $\pi = \frac{n_s}{|P_s|}$, (\textit{2}) Temporal representation via stratification (error reduction by $\sqrt{1-f}$), (\textit{3}) Cross-domain independence.
Total sample size $N = 2,400$ achieves power $\beta > 0.8$ for medium effects $(d \geq 0.5)$ at $\alpha = 0.05$ with Bonferroni correction \cite{bonferroni1936teoria}. We excluded ratings \citep{Mayzlin2012-gx, ucv047, Guo2020-gv} and helpful votes \citep{abc, Lappas2016-su, Deng2020-kl} due to documented biases. 

\subsection{EOT-X}\label{eot_x}
We introduce \textsc{\textbf{EOT-X}}, the first human-annotated benchmark dataset for emotion-opinion analysis in e-commerce. We constructed \textsc{EOT-X} by collecting annotations for $2,400$ reviews. To ensure annotation quality, each review was independently annotated by three expert raters to identify emotions (based on Plutchik’s $8$ primary emotions) and their corresponding opinion triggers. This process yielded a total of $\mathbf{7,200}$ annotations ($3$ raters × $2,400$ reviews).

\subsection{Annotation}\label{annotation_dataset}

\paragraph{Annotation Quality}  
Expert raters were selected over crowd workers due to concerns regarding annotation quality, inter-annotator agreement, and domain expertise. \cite{mohammad2013crowdsourcing} identify key challenges in crowdsourcing, such as unqualified annotators, malicious inputs, and inconsistent judgments, emphasizing the need for rigorous quality control. Given our study’s focus on nuanced emotion analysis, we employed expert raters with relevant academic training to ensure higher reliability and consistency in annotations. They followed detailed guidelines (Appendix \ref{guidelines}) and received appropriate stipends.

\begin{table}[t]
    \centering
    \resizebox{1\columnwidth}{!}{%
    \begin{tabular}{lcccc}
    \toprule
         \textbf{Domain} & \textbf{A1/A2} & \textbf{A2/A3} & \textbf{A1/A3} & \textbf{Average} \\
    \midrule
        \multicolumn{5}{c}{\textit{Emotion Agreement}}\\
    \midrule
        {\tt Beauty} & $0.88$ & $0.91$ & $0.88$ & $0.89$ \\
        {\tt Clothing} & $0.88$ & $0.90$ & $0.86$ & $0.88$ \\
        {\tt  Home} & $0.87$ & $0.89$ & $0.87$ & $0.88$ \\
        {\tt  Electronics} & $0.87$ & $0.88$ & $0.87$ & $0.87$ \\
        {\tt  TripAdvisor} & $0.85$ & $0.89$ & $0.87$ & $0.87$ \\
        {\tt Yelp} & $0.88$ & $0.91$ & $0.89$ & $0.89$ \\
        \textit{Overall Average} & $0.87$ & $0.90$ & $0.87$ & $0.88$ \\
    \midrule
        \multicolumn{5}{c}{\textit{Trigger Agreement}}\\
    \midrule
        {\tt Beauty} & $0.82$ & $0.85$ & $0.84$ & $0.84$ \\
        {\tt Clothing} & $0.81$ & $0.84$ & $0.81$ & $0.82$ \\
        {\tt  Home }& $0.81$ & $0.86$ & $0.82$ & $0.83$ \\
        {\tt  Electronics} & $0.82$ & $0.85$ & $0.83$ & $0.83$ \\
        {\tt  TripAdvisor} & $0.83$ & $0.86$ & $0.84$ & $0.85$ \\
        {\tt Yelp} & $0.83$ & $0.87$ & $0.84$ & $0.84$ \\
        \textit{Overall Average} & $0.82$ & $0.85$ & $0.83$ & $0.84$ \\
    \bottomrule
    \end{tabular}
    }
    \caption{Domain-wise annotator agreement scores for Emotion and Trigger identification.}
    \label{tab:agreement_scores}
\end{table}

\paragraph{Inter-Annotator Agreement}  
We compute inter-annotator agreement using Fleiss' Kappa~\cite{fleiss1971measuring} (\(\kappa\)) for emotions and triggers. For emotions, pairwise scores range from \(\mathbf{0.85}\) to \(\mathbf{0.91}\) across domains (Table~X), with an average of \(\mathbf{0.89}\), indicating \textit{almost perfect agreement} (\(0.81 \leq \kappa \leq 1.00\)).

For triggers, we employ a \textbf{token-level analysis} approach to better accommodate the inherent variations in span annotation. By tokenizing trigger spans, we capture partial agreements where annotators identify the same triggers with different boundaries (e.g., \texttt{"very happy"} vs. \texttt{"happy"}). Pairwise scores range from \(\mathbf{0.81}\) to \(\mathbf{0.87}\) across domains (Table~X), with an average of \(\mathbf{0.84}\), indicating \textit{almost perfect agreement}. This strong token-level agreement demonstrates consistent identification of the same key trigger elements despite span variations.

This indicates humans can reliably identify emotions and their triggers in customer reviews. Examples of EOT-X are shown in Figure \ref{fig:first}.

\begin{figure}[t]
    \centering
    \includegraphics[width=1\columnwidth]{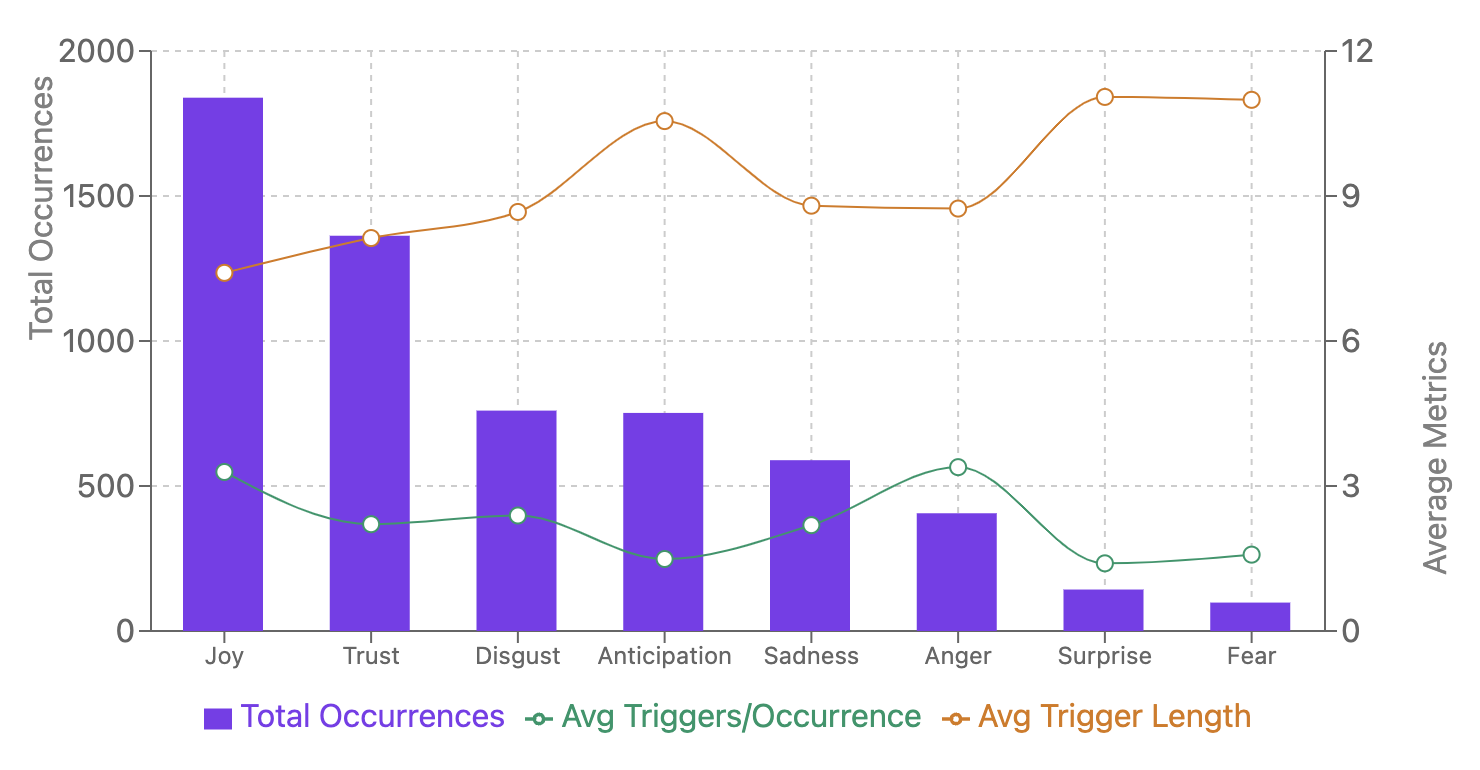}
    \caption{Overall Statistics of Gold Standard Aggregated dataset} 
    \label{fig:gold}
\end{figure}

\section{Experiments} \label{exp}




\textbf{Task-Specific Models:} Emotion-cause pair extraction (\textsc{ECPE}) systems \cite{xia-ding-2019-emotion, bao2020ecpe, ding2020ecpe, yuan2020ecpe} were designed for two-step, clause-level pair extraction in newswire texts and primarily focus on linking a single emotion clause to one cause clause. \textbf{\textit{In stark contrast}}, our work addresses a fundamentally different task—jointly detecting multiple emotions and their associated opinion triggers in e-\textit{commerce reviews}—where each emotion may be linked to several distinct textual spans.  \textit{Consequently, these methods are methodologically misaligned and irrelevant as baselines for our study}.

\textbf{Language Models:}
We conducted a comprehensive evaluation of $26$ language models, comprising $3$ Pre-trained Language Models (PLMs), $3$ proprietary models and $20$ open-source models. The full list of models and implementation details are provided in \textbf{(Appendix \ref{llms_list})} for brevity\footnote{All experiments were conducted on H100 GPUs over a period of 120+ hours.}.

\textbf{Fine-Tuned LLMs:} We fine-tuned \textls[-60]{\texttt{Llama-3.2-1B-Instruct}}, \textls[-60]{\texttt{Qwen2.5-0.5B-Instruct}}, and \textls[-60]{\texttt{DeepSeek-R1-Distill-Llama-8B}} on the \textls[-60]{\textsc{\textbf{EOT-X}}} dataset. Our objective was to assess whether compact models can match the performance of larger counterparts while reducing reliance on costly GPU infrastructure. Implementation details are provided in Appendix X.

\textbf{Rationale for Omitting PLMs for Fine-Tuning:} We chose the aforementioned LLMs over PLMs due to their superior capabilities at similar parameter counts. They support extended context (up to $128$K tokens), generate longer outputs (up to $8$K tokens), exhibit enhanced reasoning, and enable efficient fine-tuning with PEFT \cite{hu2021lora}
.

\section{Results} \label{result}

\paragraph{Evaluation on Aggregated Gold Standard Dataset}
We evaluate our models on the {\textbf{EOT-X}}. We created an \textit{aggregated gold standard} by combining annotations from three expert annotators. For emotion detection, a majority vote (at least two out of three) determines inclusion, ensuring reliability and nuance. For trigger identification, all unique triggers are preserved, and in cases of overlap the longest span is retained to capture complete context. Preprocessing steps include standardizing product titles and review texts, exact span matching, and deduplication. For statistics of aggregated gold standard refer Figure \ref{fig:gold} and Table \ref{tab:gold_standard}.

\paragraph{Model Family Behavior}
Different model families exhibit distinct characteristics. The \textsc{Llama} series shows moderate performance, while the \textsc{Mistral} family—especially the {\tt Mistral-7B-Instruct-v0.2} variants—demonstrates significant improvements with our enhanced \textsc{EOT-DETECT} framework. Similar trends are observed across the \textsc{Phi} and \textsc{Qwen} families, indicating that intrinsic family characteristics and pre-training data greatly influence performance.

\begin{table}[t]
    \centering
    \resizebox{\columnwidth}{!}{%
    \begin{tabular}{l|ccc|cccc}
    \toprule
    & \multicolumn{3}{c|}{\textsc{\textbf{Emotion}}} & \multicolumn{4}{c}{\textsc{\textbf{Opinion Trigger}}} \\
    \textbf{Models} & \textbf{P} & \textbf{R} & \textbf{F1} & \textbf{EM} & \textbf{PM} & \textbf{R1} & \textbf{RL} \\
    \midrule
    \texttt{gpt-4o\_zs}        & $0.79$   & $\underline{0.71}$   & $\underline{0.75}$   & $0.24$   & $\underline{0.54}$   & $\underline{0.73}$   & $\underline{0.62}$ \\
    \texttt{GPT-4o}\textsubscript{\scriptsize \textbf{ZS-CoT}} & $0.59$   & $0.52$   & $0.51$   & $0.26$   & $0.37$   & $0.53$   & $0.47$ \\
    \texttt{GPT-4o}\textsubscript{\scriptsize \textbf{EOT}}  & $0.69$   & $0.61$   & $0.53$   & $0.28$   & $0.38$   & $0.57$   & $0.49$ \\
    \midrule
    \texttt{o1-mini\_zs}       & $0.63$   & $0.53$   & $0.58$   & $\textbf{0.51}$   & $0.30$   & $0.68$   & $0.58$ \\
    \texttt{o1-mini\_zs Cot}   & $0.71$   & $0.59$   & $0.65$   & $\textbf{0.51}$   & $0.29$   & $0.71$   & $0.60$ \\
    \texttt{o1-mini\_EOT}      & $0.74$   & $0.66$   & $0.70$   & $\textbf{0.51}$   & $0.29$   & $0.72$   & $\underline{0.61}$ \\
    \midrule
    \texttt{claude sonet 3.5\_zs}      & $\underline{0.81}$   & $0.64$   & $0.71$   & $0.18$   & $0.52$   & $0.65$   & $0.56$ \\
    \texttt{claude sonet 3.5\_zs Cot}  & $0.69$   & $0.63$   & $0.66$   & $0.20$   & $\textbf{0.57}$   & $0.70$   & $0.59$ \\
    \texttt{claude sonet 3.5\_EOT}   & $\textbf{0.94}$   & $\textbf{0.72}$   & $\textbf{0.81}$   & $\underline{0.35}$   & $0.52$   & $\textbf{0.83}$   & $\textbf{0.68}$ \\  
    \bottomrule
    \end{tabular}
    }
    \caption{Performance comparison of closed-source models across tasks. 
    \textbf{Emotion Metrics}: P - Precision, R - Recall, F1 - F1-score. 
    \textbf{Opinion Trigger Metrics}: EM - Exact Match, PM - Partial Match, R1 - Rouge-1, RL - Rouge-L. 
    }
    \label{tab:results}
\end{table}

\paragraph{Technique Comparison}
Across both open- and closed-source settings, our \textsc{EOT-DETECT} framework consistently outperforms standard zero-shot and chain-of-thought approaches. For example, Mistral-7B-Instruct-v0.2 with \textsc{EOT-DETECT} achieves the highest emotion metrics (P: 0.86, R: 0.87, F1: 0.86), and among closed-source models, \textsc{Claude Sonnet 3.5} with \textsc{EOT-DETECT} records top scores (P: 0.94, R: 0.72, F1: 0.81). This confirms that our structured prompting framework is more effective in eliciting nuanced and complete responses from models.

\paragraph{Impact of Parameter Size}
Our results indicate a clear trend: larger models generally yield better performance. Within the \textsc{Qwen} family, increasing parameters from 0.5B to 72B leads to substantial gains in both emotion detection and trigger identification. However, improvements are also dependent on model architecture and training strategies, suggesting that model design is as crucial as scale. It is worth noting that pre-LLM approaches performed so poorly on these tasks—largely due to their limited training on token data—that we omit their results for clarity.
\paragraph{Closed-Source vs. Open-Source}
Closed-source models (e.g., GPT-4o, \textsc{Claude Sonnet 3.5}) achieve higher absolute performance across metrics. Nonetheless, the best open-source models (e.g., {\tt Mistral-7B-Instruct-v0.2 and Phi-4} with \textsc{EOT-DETECT}) perform competitively, demonstrating that state-of-the-art open-source approaches are rapidly closing the gap with commercial systems.

\begin{table}[H]
    \centering
    \resizebox{1\columnwidth}{!}{%
    \begin{tabular}{l|ccc|cccc}
    \toprule
    & \multicolumn{3}{c|}{\textsc{\textbf{Emotion}}} & \multicolumn{4}{c}{\textsc{\textbf{Opinion Trigger}}} \\
    \textbf{Models} & \textbf{P} & \textbf{R} & \textbf{F1} & \textbf{EM} & \textbf{PM} & \textbf{R1} & \textbf{RL} \\
    \midrule
    \texttt{Llama-3.2-1B-Instruct\_zs}           & $0.13$   & $0.20$   & $0.10$   & $0.03$   & $0.10$   & $0.01$   & $0.01$ \\
    \texttt{Llama-3.2-1B-Instruct\_zs Cot}       & $0.16$   & $0.31$   & $0.11$   & $0.09$   & $0.09$   & $0.10$   & $0.10$ \\
    \texttt{Llama-3.2-1B-Instruct\_EOT}          & $0.23$   & $0.57$   & $0.33$   & $0.11$   & $0.13$   & $0.33$   & $0.28$ \\
    \midrule
    \texttt{Llama-3.2-3B-Instruct\_zs}           & $0.17$   & $0.07$   & $0.10$   & $0.04$   & $0.04$   & $0.05$   & $0.05$ \\
    \texttt{Llama-3.2-3B-Instruct\_zs Cot}       & $0.21$   & $0.19$   & $0.16$   & $0.11$   & $0.09$   & $0.019$   & $0.17$ \\
    \texttt{Llama-3.2-3B-Instruct\_EOT}          & $0.49$   & $0.39$   & $0.44$   & $0.26$   & $0.31$   & $0.42$   & $0.37$ \\
    \midrule
    \texttt{Llama-3.1-8B-Instruct\_zs}           & $0.59$   & $0.52$   & $0.51$   & $0.26$   & $0.37$   & $0.53$   & $0.47$ \\
    \texttt{Llama-3.1-8B-Instruct\_zs Cot}       & $0.10$   & $0.08$   & $0.06$   & $0.10$   & $0.10$   & $0.09$   & $0.05$ \\
    \texttt{Llama-3.1-8B-Instruct\_EOT}          & $0.18$   & $0.16$   & $0.17$   & $0.10$   & $0.10$   & $0.17$   & $0.14$ \\
    \midrule
    \texttt{Llama-3.3-70B-Instruct\_zs}          & $0.19$   & $0.16$   & $0.18$   & $0.06$   & $0.13$   & $0.17$   & $0.14$ \\
    \texttt{Llama-3.3-70B-Instruct\_zs Cot}      & $0.14$   & $0.11$   & $0.08$   & $0.03$   & $0.03$   & $0.07$   & $0.04$ \\
    \texttt{Llama-3.3-70B-Instruct\_EOT}         & $0.18$   & $0.21$   & $0.19$   & $0.06$   & $0.12$   & $0.18$   & $0.15$ \\
    \midrule
    \texttt{Mistral-7B-Instruct-v0.2\_zs}        & $0.69$   & $0.45$   & $0.54$   & $0.41$   & $0.33$   & $0.59$   & $0.50$ \\
    \texttt{Mistral-7B-Instruct-v0.2\_zs Cot}    & $0.71$   & $0.63$   & $0.69$   & $0.41$   & $0.33$   & $0.39$   & $0.48$ \\
    \texttt{Mistral-7B-Instruct-v0.2\_EOT}       & $\textbf{0.86}$   & $\textbf{0.87}$   & $\textbf{0.86}$   & $\textbf{0.55}$   & $0.34$   & $\textbf{0.82}$   & $\textbf{0.68}$ \\
    \midrule
    \texttt{Mistral-7B-Instruct-v0.3\_zs}        & $0.34$   & $0.18$   & $0.23$   & $0.22$   & $0.13$   & $0.25$   & $0.22$ \\
    \texttt{Mistral-7B-Instruct-v0.3\_zs Cot}    & $0.38$   & $0.30$   & $0.34$   & $0.16$   & $0.41$   & $0.58$   & $0.50$ \\
    \texttt{Mistral-7B-Instruct-v0.3\_EOT}       & $0.59$   & $0.54$   & $0.57$   & $0.30$   & $0.32$   & $0.60$   & $0.50$ \\
    \midrule
    \texttt{Mistral-Small-24B-Instruct-2501\_zs}       & $0.38$   & $0.37$   & $0.37$   & $0.15$   & $0.27$   & $0.38$   & $0.30$ \\
    \texttt{Mistral-Small-24B-Instruct-2501\_zs Cot}   & $0.33$   & $0.36$   & $0.34$   & $0.10$   & $0.26$   & $0.35$   & $0.28$ \\
    \texttt{Mistral-Small-24B-Instruct-2501\_EOT}      & $0.37$   & $0.38$   & $0.38$   & $0.13$   & $0.26$   & $0.38$   & $0.30$ \\
    \midrule
    \texttt{zephyr-7b-beta\_zs}               & $0.52$   & $0.22$   & $0.31$   & $0.37$   & $0.17$   & $0.40$   & $0.35$ \\
    \texttt{zephyr-7b-beta\_zs Cot}           & $0.59$   & $0.54$   & $0.57$   & $0.30$   & $0.32$   & $0.60$   & $0.50$ \\
    \texttt{zephyr-7b-beta\_EOT}              & $0.66$   & $0.30$   & $0.41$   & $0.26$   & $0.17$   & $0.40$   & $0.33$ \\
    \midrule
    \texttt{Phi-3.5-mini-instruct\_zs}        & $0.63$   & $0.53$   & $0.58$   & $0.51$   & $0.30$   & $0.68$   & $0.58$ \\
    \texttt{Phi-3.5-mini-instruct\_zs Cot}    & $0.71$   & $0.59$   & $0.65$   & $0.51$   & $0.29$   & $0.71$   & $0.60$ \\
    \texttt{Phi-3.5-mini-instruct\_EOT}       & $0.74$   & $0.66$   & $0.70$   & $0.51$   & $0.29$   & $0.72$   & $0.61$ \\
    \midrule
    \texttt{Phi-4\_zs}                        & $0.81$   & $0.68$   & $0.74$   & $0.49$   & $0.38$   & $0.78$   & $0.65$ \\
    \texttt{Phi-4\_zs Cot}                    & $0.70$   & $0.69$   & $0.70$   & $0.40$   & $0.41$   & $0.73$   & $0.60$ \\
    \texttt{Phi-4\_EOT}                       & $\textbf{0.86}$   & $\textbf{0.87}$   & $\textbf{0.86}$   & $0.54$   & $0.35$   & $0.82$   & $0.67$ \\
    \midrule
    \texttt{gemma-2-2b-it\_zs}           & $0.72$   & $0.44$   & $0.55$   & $0.38$   & $0.39$   & $0.44$   & $0.40$ \\
    \texttt{gemma-2-2b-it\_zs Cot}       & $0.70$   & $0.69$   & $0.70$   & $0.51$   & $0.29$   & $0.71$   & $0.60$ \\
    \texttt{gemma-2-2b-it\_EOT}          & $0.75$   & $0.49$   & $0.59$   & $0.27$   & $0.42$   & $0.52$   & $0.46$ \\
    \midrule
    \texttt{gemma-2-9b-it\_zs}           & $0.72$   & $0.54$   & $0.61$   & $0.32$   & $0.47$   & $0.64$   & $0.56$ \\
    \texttt{gemma-2-9b-it\_zs Cot}       & $0.34$   & $0.18$   & $0.23$   & $0.22$   & $0.13$   & $0.25$   & $0.22$ \\
    \texttt{gemma-2-9b-it\_EOT}          & $0.56$   & $0.39$   & $0.44$   & $0.21$   & $0.36$   & $0.47$   & $0.41$ \\
    \midrule
    \texttt{gemma-2-27b-it\_zs}          & $0.49$   & $0.38$   & $0.43$   & $0.31$   & $0.28$   & $0.45$   & $0.38$ \\
    \texttt{gemma-2-27b-it\_zs Cot}      & $0.33$   & $0.36$   & $0.34$   & $0.10$   & $0.26$   & $0.35$   & $0.28$ \\
    \texttt{gemma-2-27b-it\_EOT}         & $0.34$   & $0.29$   & $0.31$   & $0.18$   & $0.21$   & $0.31$   & $0.26$ \\
    \midrule
    \texttt{Qwen2.5-0.5B-Instruct\_zs}      & $0.38$   & $0.30$   & $0.34$   & $0.16$   & $0.41$   & $0.58$   & $0.50$ \\
    \texttt{Qwen2.5-0.5B-Instruct\_zs Cot}  & $0.35$   & $0.37$   & $0.36$   & $0.12$   & $0.36$   & $0.55$   & $0.48$ \\
    \texttt{Qwen2.5-0.5B-Instruct\_EOT}     & $0.30$   & $0.46$   & $0.36$   & $0.12$   & $0.31$   & $0.52$   & $0.45$ \\
    \midrule
    \texttt{Qwen2.5-3B-Instruct\_zs}         & $0.63$   & $0.38$   & $0.48$   & $0.26$   & $0.30$   & $0.35$   & $0.32$ \\
    \texttt{Qwen2.5-3B-Instruct\_zs Cot}     & $0.56$   & $0.67$   & $0.61$   & $0.31$   & $0.30$   & $0.56$   & $0.48$ \\
    \texttt{Qwen2.5-3B-Instruct\_EOT}        & $0.63$   & $0.43$   & $0.51$   & $0.29$   & $0.35$   & $0.48$   & $0.42$ \\
    \midrule
    \texttt{Qwen2.5-7B-Instruct\_zs}         & $0.75$   & $0.44$   & $0.56$   & $0.26$   & $0.48$   & $0.54$   & $0.47$ \\
    \texttt{Qwen2.5-7B-Instruct\_zs Cot}     & $0.61$   & $0.62$   & $0.61$   & $0.28$   & $0.37$   & $0.58$   & $0.49$ \\
    \texttt{Qwen2.5-7B-Instruct\_EOT}        & $0.67$   & $0.46$   & $0.55$   & $0.26$   & $0.44$   & $0.54$   & $0.47$ \\
    \midrule
    \texttt{Qwen2.5-32B-Instruct\_zs}        & $0.78$   & $0.54$   & $0.63$   & $0.33$   & $\textbf{0.51}$   & $0.72$   & $0.60$ \\
    \texttt{Qwen2.5-32B-Instruct\_zs Cot}    & $0.71$   & $0.66$   & $0.69$   & $0.31$   & $0.48$   & $0.71$   & $0.59$ \\
    \texttt{Qwen2.5-32B-Instruct\_EOT}       & $\textbf{0.80}$   & $\underline{0.80}$   & $\underline{0.80}$   & $0.42$   & $0.43$   & $0.78$   & $0.64$ \\
    \midrule
    \texttt{Qwen2.5-72B-Instruct\_zs}        & $\underline{0.85}$   & $0.61$   & $0.71$   & $0.33$   & $0.51$   & $0.72$   & $0.60$ \\
    \texttt{Qwen2.5-72B-Instruct\_zs Cot}    & $0.71$   & $0.71$   & $0.71$   & $0.38$   & $0.41$   & $0.74$   & $0.61$ \\
    \texttt{Qwen2.5-72B-Instruct\_EOT}       & $0.76$   & $0.73$   & $0.75$   & $0.30$   & $0.51$   & $0.76$   & $0.62$ \\
    \midrule
    \texttt{QwQ-32B-Preview\_zs}             & $0.63$   & $0.61$   & $0.61$   & $0.35$   & $0.35$   & $0.69$   & $0.56$ \\
    \texttt{QwQ-32B-Preview\_zs Cot}         & $0.62$   & $0.59$   & $0.61$   & $0.28$   & $0.38$   & $0.66$   & $0.55$ \\
    \texttt{QwQ-32B-Preview\_EOT}            & $0.64$   & $0.73$   & $0.68$   & $0.32$   & $0.40$   & $0.72$   & $0.59$ \\
    \midrule
    \texttt{Deepseek-R1\_zs}                  & $0.78$   & $0.66$   & $0.71$   & $0.32$   & $0.49$   & $0.71$   & $0.60$ \\
    \texttt{Deepseek-R1\_zs Cot}              & $0.75$   & $0.70$   & $0.72$   & $0.27$   & $0.51$   & $0.70$   & $0.59$ \\
    \texttt{Deepseek-R1\_EOT}                 & $0.75$   & $0.73$   & $0.74$   & $0.28$   & $0.45$   & $0.72$   & $0.60$ \\
    \bottomrule
    \end{tabular}
    }
   \caption{Performance comparison of open-source models. \textbf{Emotion Metrics}: P (Precision), R (Recall), F1 (F1-score). \textbf{Opinion Trigger Metrics}: EM (Exact Match), PM (Partial Match), R1 (Rouge-1), RL (Rouge-L). \textbf{ZS}, \textbf{ZS-CoT}, \textbf{EOT-DETECT}}
    \label{tab:closed_results}
\end{table}

\paragraph{Fine-Tuned Models:}
Additionally, fine-tuned variants (e.g., Llama-3.2-1B-Instruct-FT (EOT-LlaMa)) generally show improved consistency over non-fine-tuned counterparts. While fine-tuning provides task-specific gains, our prompting strategies—particularly EOT—remain robust and effective across diverse domains.
\begin{table}[H]
    \centering
    \resizebox{\columnwidth}{!}{%
    \begin{tabular}{l|ccccccc}
    \toprule
    \textbf{Models} & \textbf{P} & \textbf{R} & \textbf{F1} & \textbf{EM} & \textbf{PM} & \textbf{R1} & \textbf{RL} \\
    \midrule
    \textbf{Qwen2.5-0.5B-Instruct-FT}                & 0.33 & 0.32 & 0.32 & 0.01 & 0.05 & 0.20 & 0.15 \\
    \textbf{Llama-3.2-1B-Instruct-FT (EOT-LlaMa)}     & 0.45 & 0.52 & 0.49 & 0.20 & 0.33 & 0.43 & 0.37 \\
    \textbf{DeepSeek-R1-Distill-Llama-8B}            & 0.36 & 0.33 & 0.35 & 0.02 & 0.04 & 0.20 & 0.16 \\
    \bottomrule
    \end{tabular}
    }
    \caption{Performance metrics for Finetuned models}
    \label{tab:ft_models}
\end{table}
\paragraph{Error Analysis}
Through detailed examination of model outputs, we identified several common failure patterns:

\begin{itemize}
    \item \textbf{Trigger Boundary Detection:} Models often struggle with precise trigger span boundaries, particularly for complex emotional expressions. For instance, in cases where emotions are expressed through multiple connected clauses, models sometimes over-extend or under-extend the trigger spans.
    \item \textbf{Emotion Conflation:} We observed that models occasionally conflate similar emotions, particularly within Plutchik's adjacent categories. For example, ``Trust'' and ``Joy'' are frequently confused when the review expresses satisfaction with product reliability.
    \item \textbf{Context Sensitivity:} Performance degrades noticeably when emotions are expressed through implicit or contextual cues rather than explicit statements. This is particularly evident in reviews containing sarcasm or subtle emotional undertones.
    \item \textbf{Trigger Multiplicity:} Models show inconsistent performance when handling multiple triggers for a single emotion, often missing secondary or tertiary triggers that human annotators readily identify.
\end{itemize}

\paragraph{Summary}
Our comprehensive evaluation—based on a rigorously aggregated gold standard—shows that:
(\textit{1}) Model families differ markedly in their baseline and enhanced performance.
(\textit{2}) The \textsc{EOT-DETECT} framework delivers consistent and significant gains over standard methods.
(\textit{3} Larger parameter sizes generally translate to improved performance, with architectural design playing a critical role.
(\textit{4}) Although closed-source models currently hold a performance edge, open-source models are rapidly catching up.

These findings underscore the importance of structured prompt engineering, model scale, and aggregation methodologies in advancing the state of emotion and trigger detection in natural language processing.

\section{Conclusion \& Future Work}
This work introduced \textsc{\textbf{EOT}} (\textsc{Emotion-Opinion Trigger}), the first joint task that unifies emotion detection and opinion trigger extraction in e-commerce reviews. We demonstrated why a focused set of Plutchik's $8$ primary emotions provides the optimal framework for analyzing customer feedback, supported by empirical evidence of strong inter-annotator agreement (emotion $\kappa=0.89$, trigger $\kappa=0.84$).
Some key takeaways are: (\textit{a}) Our work presents the \textsc{\textbf{EOT-X}} benchmark—a rigorously annotated dataset. (\textit{b}) \textsc{\textbf{EOT-DETECT}}, a structured prompting framework that achieves superior performance over conventional zero-shot and chain-of-thought methods. (\textit{c}) \texttt{\textbf{EOT-LLaMA}}, an edge-deployable $1$B-parameter model that outperforms larger models while maintaining efficiency.

Our work aims to establish a foundation for understanding not just \textit{what} emotions customers express, but \textit{why} they feel them, enabling more nuanced analysis of user feedback. We hope our work catalyzes further advances in emotion-opinion analysis systems for e-commerce applications.

\section*{Limitations}

We could not test state-of-the-art OpenAI's reasoning models \textls[-60]{\tt o1, o3-mini} and \textls[-60]{\tt o3-mini-high} as these required higher-tier API access beyond our research budget. We observed that when prompted, Open-AI models {\tt GPT-4o} and {\tt o1-mini} prevented access to their internal chain-of-thought (CoT) reasoning steps. This opacity limited our ability to analyze how these models interacted with our framework and hinders further improvements. Note that this is due to the proprietary, closed-source nature of these AI models rather a limitation of our framework. While we believe the development of the \textsc{\textbf{EOT-X}} dataset offers new research opportunities in emotion-opinion analysis, its moderate size of $\mathbf{2,400}$ samples stems from the complexity of annotating emotions and opinion triggers at the phrase level, which is both labor-intensive and financially demanding in academic research.

\section*{Ethical Considerations}
We engaged three raters with diverse academic backgrounds: a Master's student, a Pre-Doctoral researcher, and a Doctoral candidate. All were male, aged $24$–$32$, with publications or active research in emotion analysis and coursework in Consumer Psychology. Raters were compensated appropriately for their contributions.

\textls[-60]{\textsc{\textbf{EOT-DETECT}}} framework and the \textsc{\tt{EOT-LLaMA}} model, a $1$B-parameter model fine-tuned for emotion-trigger analysis in customer reviews—may occasionally produce hallucinations or unintended outputs, particularly in complex cases.

In this work, we define emotion detection and opinion trigger extraction within the context of analyzing customer reviews to understand the interplay between expressed emotions and their underlying causes.
We do not claim that LLMs can experience or replicate emotions. Researchers should verify framework and model appropriateness before integration into evaluation processes, ensuring careful application in real-world e-commerce scenarios.
\bibliography{anthology,custom}

\begin{thebibliography}{58}
\expandafter\ifx\csname natexlab\endcsname\relax\def\natexlab#1{#1}\fi

\bibitem[{Acheampong et~al.(2023)Acheampong, Nunoo-Mensah, and Chen}]{acheampong2023text}
Francisca~Adoma Acheampong, Henry Nunoo-Mensah, and Wei Chen. 2023.
\newblock \href {https://doi.org/10.1002/eng2.12549} {Text-based emotion detection: Advances, challenges, and opportunities}.
\newblock \emph{Engineering Reports}, 5(4):e12549.

\bibitem[{AI@Meta(2024)}]{l3}
AI@Meta. 2024.
\newblock \href {https://github.com/meta-llama/llama3/blob/main/MODEL_CARD.md} {Llama 3 model card}.

\bibitem[{Bao et~al.(2020)Bao, Zhang, and Liu}]{bao2020ecpe}
Yujie Bao, Min Zhang, and Yang Liu. 2020.
\newblock \href {https://doi.org/10.18653/v1/2020.acl-main.91} {Emotion-cause pair extraction with graph convolutional networks}.
\newblock In \emph{Proceedings of the 58th Annual Meeting of the Association for Computational Linguistics}, pages 1010--1020, Online. Association for Computational Linguistics.

\bibitem[{Bian and Yan(2022)}]{bian2022analyzing}
Weijun Bian and Gong Yan. 2022.
\newblock \href {https://doi.org/10.3389/fpsyg.2022.884673} {Analyzing intention to purchase brand extension via brand attribute associations: The mediating and moderating role of emotional consumer-brand relationship and brand commitment}.
\newblock \emph{Frontiers in Psychology}, 13:884673.

\bibitem[{Bonferroni(1936)}]{bonferroni1936teoria}
Corrado Bonferroni. 1936.
\newblock Teoria statistica delle classi e calcolo delle probabilità.
\newblock \emph{Pubblicazioni del R Istituto Superiore di Scienze Economiche e Commerciali di Firenze}, 8:3--62.

\bibitem[{Brown et~al.(2020)Brown, Mann, Ryder, Subbiah, Kaplan, Dhariwal, Neelakantan, Shyam, Sastry, Askell, Agarwal, Herbert-Voss, Krueger, Henighan, Child, Ramesh, Ziegler, Wu, Winter, Hesse, Chen, Sigler, Litwin, Gray, Chess, Clark, Berner, McCandlish, Radford, Sutskever, and Amodei}]{brown2020language}
Tom~B. Brown, Benjamin Mann, Nick Ryder, Melanie Subbiah, Jared Kaplan, Prafulla Dhariwal, Arvind Neelakantan, Pranav Shyam, Girish Sastry, Amanda Askell, Sandhini Agarwal, Ariel Herbert-Voss, Gretchen Krueger, Tom Henighan, Rewon Child, Aditya Ramesh, Daniel~M. Ziegler, Jeffrey Wu, Clemens Winter, Christopher Hesse, Mark Chen, Eric Sigler, Mateusz Litwin, Scott Gray, Benjamin Chess, Jack Clark, Christopher Berner, Sam McCandlish, Alec Radford, Ilya Sutskever, and Dario Amodei. 2020.
\newblock \href {https://proceedings.neurips.cc/paper/2020/file/1457c0d6bfcb4967418bfb8ac142f64a-Paper.pdf} {Language models are few-shot learners}.
\newblock In \emph{Advances in Neural Information Processing Systems}, volume~33, pages 1877--1901.

\bibitem[{Chen et~al.(2022)Chen, Samaranayake, Cen, Qi, and Lan}]{chen2022impact}
Tao Chen, Premaratne Samaranayake, XiongYing Cen, Meng Qi, and Yi-Chen Lan. 2022.
\newblock The impact of online reviews on consumers’ purchasing decisions: Evidence from an eye-tracking study.
\newblock \emph{Frontiers in Psychology}, 13:865702.

\bibitem[{Cochran(1977)}]{cochran1977sampling}
William~G. Cochran. 1977.
\newblock \emph{Sampling Techniques}, 3rd edition.
\newblock Wiley.

\bibitem[{de~Langhe et~al.(2015)de~Langhe, Fernbach, and Lichtenstein}]{ucv047}
Bart de~Langhe, Philip~M. Fernbach, and Donald~R. Lichtenstein. 2015.
\newblock \href {https://doi.org/10.1093/jcr/ucv047} {{Navigating by the Stars: Investigating the Actual and Perceived Validity of Online User Ratings}}.
\newblock \emph{Journal of Consumer Research}, 42(6):817--833.

\bibitem[{Demszky et~al.(2020)Demszky, Movshovitz-Attias, Ko, Cowen, Nemade, and Ravi}]{demszky-etal-2020-goemotions}
Dorottya Demszky, Dana Movshovitz-Attias, Jeongwoo Ko, Alan Cowen, Gaurav Nemade, and Sujith Ravi. 2020.
\newblock \href {https://doi.org/10.18653/v1/2020.acl-main.372} {{G}o{E}motions: A dataset of fine-grained emotions}.
\newblock In \emph{Proceedings of the 58th Annual Meeting of the Association for Computational Linguistics}, pages 4040--4054, Online. Association for Computational Linguistics.

\bibitem[{Deng et~al.(2020)Deng, Yi, and Lu}]{Deng2020-kl}
Weihua Deng, Ming Yi, and Yingying Lu. 2020.
\newblock Vote or not? how various information cues affect helpfulness voting of online reviews.
\newblock \emph{Online Inf. Rev.}, 44(4):787--803.

\bibitem[{Ding et~al.(2020)Ding, Xia, and Yu}]{ding2020ecpe}
Zixiang Ding, Rui Xia, and Jianfei Yu. 2020.
\newblock Ecpe-2d: Emotion-cause pair extraction based on joint two-dimensional representation, interaction and prediction.
\newblock In \emph{Proceedings of the 58th Annual Meeting of the Association for Computational Linguistics}, pages 3161--3170.

\bibitem[{Ekman(1992)}]{ekman1992argument}
Paul Ekman. 1992.
\newblock \href {https://doi.org/10.1080/02699939208411068} {An argument for basic emotions}.
\newblock \emph{Cognition and Emotion}, 6(3-4):169--200.

\bibitem[{Fan et~al.(2021)Fan, Li, Li, Yan, and Wu}]{fan-etal-2021-joint}
Wenqian Fan, Pengfei Li, Wei Li, Rui Yan, and Lide Wu. 2021.
\newblock \href {https://doi.org/10.18653/v1/2021.acl-long.62} {Joint constrained learning with boundary-adjusting for emotion-cause pair extraction}.
\newblock In \emph{Proceedings of the 59th Annual Meeting of the Association for Computational Linguistics}, pages 1570--1580, Online. Association for Computational Linguistics.

\bibitem[{Fiske and Taylor(1991)}]{fiske1991social}
Susan~T. Fiske and Shelley~E. Taylor. 1991.
\newblock \emph{Social Cognition}, 2nd edition.
\newblock McGraw-Hill, New York.

\bibitem[{Fleiss(1971)}]{fleiss1971measuring}
Jerome~L. Fleiss. 1971.
\newblock \emph{Measuring Nominal Scale Agreement Among Many Raters}, volume~76.
\newblock American Psychological Association.

\bibitem[{Greifeneder et~al.(2007)Greifeneder, Bless, and Kuschmann}]{greifeneder2007extending}
Rainer Greifeneder, Herbert Bless, and Thorsten Kuschmann. 2007.
\newblock Extending the brand image on new products: The facilitative effect of happy mood states.
\newblock \emph{Journal of Consumer Behaviour: An International Research Review}, 6(1):19--31.

\bibitem[{Gui et~al.(2016)Gui, Wu, Xu, Lu, and Zhou}]{gui-etal-2016-event}
Lin Gui, Dongyin Wu, Ruifeng Xu, Qin Lu, and Yu~Zhou. 2016.
\newblock \href {https://doi.org/10.18653/v1/D16-1170} {Event-driven emotion cause extraction with corpus construction}.
\newblock In \emph{Proceedings of the 2016 Conference on Empirical Methods in Natural Language Processing}, pages 1639--1649, Austin, Texas. Association for Computational Linguistics.

\bibitem[{Guo et~al.(2020)Guo, Wang, and Wu}]{Guo2020-gv}
Junpeng Guo, Xiaopan Wang, and Yi~Wu. 2020.
\newblock Positive emotion bias: Role of emotional content from online customer reviews in purchase decisions.
\newblock \emph{J. Retail. Consum. Serv.}, 52(101891):101891.

\bibitem[{Herrando and Constantinides(2021)}]{Herrando2021-pm}
Carolina Herrando and Efthymios Constantinides. 2021.
\newblock Emotional contagion: A brief overview and future directions.
\newblock \emph{Front. Psychol.}, 12:712606.

\bibitem[{Hou et~al.(2024)Hou, Li, He, Yan, Chen, and McAuley}]{hou2024bridging}
Yupeng Hou, Jiacheng Li, Zhankui He, An~Yan, Xiusi Chen, and Julian McAuley. 2024.
\newblock \href {https://arxiv.org/abs/2403.03952} {Bridging language and items for retrieval and recommendation}.
\newblock \emph{arXiv preprint arXiv:2403.03952}.

\bibitem[{Hu et~al.(2021)Hu, Shen, Wallis, Allen-Zhu, Li, Wang, Wang, and Chen}]{hu2021lora}
Edward~J. Hu, Yelong Shen, Phillip Wallis, Zeyuan Allen-Zhu, Yuanzhi Li, Shean Wang, Lu~Wang, and Weizhu Chen. 2021.
\newblock \href {https://arxiv.org/abs/2106.09685} {Lora: Low-rank adaptation of large language models}.
\newblock \emph{arXiv preprint arXiv:2106.09685}.

\bibitem[{Huang and Rust(2024)}]{huang2024feeling}
Ming-Hui Huang and Roland~T. Rust. 2024.
\newblock \href {https://doi.org/10.1177/00222429231224748} {Feeling ai for customer care}.
\newblock \emph{Journal of Marketing}, 88(1):23--44.

\bibitem[{Jiang et~al.(2023)Jiang, Sablayrolles, Mensch, Bamford, Chaplot, de~las Casas, Bressand, Lengyel, Lample, Saulnier, Lavaud, Lachaux, Stock, Scao, Lavril, Wang, Lacroix, and Sayed}]{m2}
Albert~Q. Jiang, Alexandre Sablayrolles, Arthur Mensch, Chris Bamford, Devendra~Singh Chaplot, Diego de~las Casas, Florian Bressand, Gianna Lengyel, Guillaume Lample, Lucile Saulnier, Lélio~Renard Lavaud, Marie-Anne Lachaux, Pierre Stock, Teven~Le Scao, Thibaut Lavril, Thomas Wang, Timothée Lacroix, and William~El Sayed. 2023.
\newblock \href {http://arxiv.org/abs/2310.06825} {Mistral 7b}.

\bibitem[{Jiang et~al.(2024)Jiang, Sablayrolles, Roux, Mensch, Savary, Bamford, Chaplot, de~las Casas, Hanna, Bressand, Lengyel, Bour, Lample, Lavaud, Saulnier, Lachaux, Stock, Subramanian, Yang, Antoniak, Scao, Gervet, Lavril, Wang, Lacroix, and Sayed}]{mix}
Albert~Q. Jiang, Alexandre Sablayrolles, Antoine Roux, Arthur Mensch, Blanche Savary, Chris Bamford, Devendra~Singh Chaplot, Diego de~las Casas, Emma~Bou Hanna, Florian Bressand, Gianna Lengyel, Guillaume Bour, Guillaume Lample, Lélio~Renard Lavaud, Lucile Saulnier, Marie-Anne Lachaux, Pierre Stock, Sandeep Subramanian, Sophia Yang, Szymon Antoniak, Teven~Le Scao, Théophile Gervet, Thibaut Lavril, Thomas Wang, Timothée Lacroix, and William~El Sayed. 2024.
\newblock \href {http://arxiv.org/abs/2401.04088} {Mixtral of experts}.

\bibitem[{Kim et~al.(2019)Kim, Park, and Park}]{kim2019emotion}
Sang-Hyun Kim, Kyung~Hoon Park, and Jiyoung Park. 2019.
\newblock \href {https://doi.org/10.1108/INTR-09-2018-0415} {Emotion as a signal of product quality: Its effect on purchase decision}.
\newblock \emph{Internet Research}, 29(5):1103--1124.

\bibitem[{Kim and Sullivan(2019)}]{Kim2019-gs}
Youn-Kyung Kim and Pauline Sullivan. 2019.
\newblock Emotional branding speaks to consumers' heart: the case of fashion brands.
\newblock \emph{Fashion Text.}, 6(1).

\bibitem[{Kojima et~al.(2022)Kojima, Gu, Reid, Matsuo, and Iwasawa}]{kojima2022large}
Takeshi Kojima, Shixiang~Shane Gu, Machel Reid, Yutaka Matsuo, and Yusuke Iwasawa. 2022.
\newblock \href {https://arxiv.org/abs/2205.11916} {Large language models are zero-shot reasoners}.
\newblock \emph{arXiv preprint arXiv:2205.11916}.

\bibitem[{Lappas et~al.(2016)Lappas, Sabnis, and Valkanas}]{Lappas2016-su}
Theodoros Lappas, Gaurav Sabnis, and Georgios Valkanas. 2016.
\newblock The impact of fake reviews on online visibility: A vulnerability assessment of the hotel industry.
\newblock \emph{Inf. Syst. Res.}, 27(4):940--961.

\bibitem[{Lee et~al.(2010)Lee, Chen, and Huang}]{lee-etal-2010-text}
Sophia Yat~Mei Lee, Ying Chen, and Chu-Ren Huang. 2010.
\newblock \href {https://aclanthology.org/W10-0206} {A text-driven rule-based system for emotion cause detection}.
\newblock In \emph{Proceedings of the {NAACL} {HLT} 2010 Workshop on Computational Approaches to Analysis and Generation of Emotion in Text}, pages 45--53, Los Angeles, CA. Association for Computational Linguistics.

\bibitem[{Li et~al.(2014)Li, Ott, Cardie, and Hovy}]{li-etal-2014-towards}
Jiwei Li, Myle Ott, Claire Cardie, and Eduard Hovy. 2014.
\newblock \href {https://doi.org/10.3115/v1/P14-1147} {Towards a general rule for identifying deceptive opinion spam}.
\newblock In \emph{Proceedings of the 52nd Annual Meeting of the Association for Computational Linguistics (Volume 1: Long Papers)}, pages 1566--1576, Baltimore, Maryland. Association for Computational Linguistics.

\bibitem[{Li et~al.(2019)Li, Feng, Wang, and Zhang}]{li2019context}
Xiangju Li, Shi Feng, Daling Wang, and Yifei Zhang. 2019.
\newblock \href {https://doi.org/10.1016/j.knosys.2019.03.008} {Context-aware emotion cause analysis with multi-attention-based neural network}.
\newblock \emph{Knowledge-Based Systems}, 174:205--218.

\bibitem[{Malik and Bilal(2024)}]{malik2024nlp}
Nadia Malik and Muhammad Bilal. 2024.
\newblock \href {https://doi.org/10.7717/peerj-cs.2203} {Natural language processing for analyzing online customer reviews: a survey, taxonomy, and open research challenges}.
\newblock \emph{PeerJ Computer Science}, 10:e2203.

\bibitem[{Mayzlin et~al.(2012)Mayzlin, Dover, and Chevalier}]{Mayzlin2012-gx}
Dina Mayzlin, Yaniv Dover, and Judith~A Chevalier. 2012.
\newblock Promotional reviews: An empirical investigation of online review manipulation.
\newblock \emph{SSRN Electron. J.}

\bibitem[{Mohammad and Turney(2013)}]{mohammad2013crowdsourcing}
Saif~M. Mohammad and Peter~D. Turney. 2013.
\newblock \href {https://doi.org/10.1111/j.1467-8640.2012.00460.x} {Crowdsourcing a word--emotion association lexicon}.
\newblock \emph{Computational Intelligence}, 29(3):436--465.

\bibitem[{Neviarouskaya et~al.(2009)Neviarouskaya, Prendinger, and Ishizuka}]{neviarouskaya2009compositionality}
Alena Neviarouskaya, Helmut Prendinger, and Mitsuru Ishizuka. 2009.
\newblock \href {https://www.aaai.org/ocs/index.php/ICWSM/09/paper/view/154} {Compositionality principle in recognition of fine-grained emotions from text}.
\newblock In \emph{Proceedings of the International Conference on Weblogs and Social Media (ICWSM)}, pages 278--285. Association for the Advancement of Artificial Intelligence (AAAI).

\bibitem[{OpenAI(2023)}]{openai}
OpenAI. 2023.
\newblock \href {https://arxiv.org/abs/2303.08774} {Gpt-4 technical report}.
\newblock \emph{ArXiv}, abs/2303.08774.

\bibitem[{Ouyang et~al.(2022)Ouyang, Wu, Jiang, Almeida, Wainwright, Mishkin, Zhang, Agarwal, Slama, Ray, Schulman, Hilton, Kelton, Miller, Simens, Askell, Welinder, Christiano, Leike, and Lowe}]{ouyang2022training}
Long Ouyang, Jeff Wu, Xu~Jiang, Diogo Almeida, Carroll~L. Wainwright, Pamela Mishkin, Chong Zhang, Sandhini Agarwal, Katarina Slama, Alex Ray, John Schulman, Jacob Hilton, Fraser Kelton, Luke Miller, Maddie Simens, Amanda Askell, Peter Welinder, Paul Christiano, Jan Leike, and Ryan Lowe. 2022.
\newblock \href {https://proceedings.neurips.cc/paper_files/paper/2022/file/b1efde53be364a73914f58805a001731-Paper-Conference.pdf} {Training language models to follow instructions with human feedback}.
\newblock In \emph{Advances in Neural Information Processing Systems}, volume~35, pages 27730--27744.

\bibitem[{Pham(2007)}]{pham2007emotion}
Michel~Tuan Pham. 2007.
\newblock Emotion and rationality: A critical review and interpretation of empirical evidence.
\newblock \emph{Review of general psychology}, 11(2):155--178.

\bibitem[{Plutchik(1980)}]{plutchik1980emotion}
Robert Plutchik. 1980.
\newblock \emph{Emotion: A Psychoevolutionary Synthesis}.
\newblock Harper \& Row, New York.

\bibitem[{Plutchik(2001)}]{plutchik2001nature}
Robert Plutchik. 2001.
\newblock \href {https://doi.org/10.1511/2001.4.344} {The nature of emotions}.
\newblock \emph{American Scientist}, 89(4):344--350.

\bibitem[{Rosario et~al.(2016)Rosario, Sotgiu, Valck, and Bijmolt}]{baba}
Ana~Babić Rosario, Francesca Sotgiu, Kristine~De Valck, and Tammo~H.A. Bijmolt. 2016.
\newblock \href {https://doi.org/10.1509/jmr.14.0380} {The effect of electronic word of mouth on sales: A meta-analytic review of platform, product, and metric factors}.
\newblock \emph{Journal of Marketing Research}, 53(3):297--318.

\bibitem[{Russell(1980)}]{russell1980circumplex}
James~A. Russell. 1980.
\newblock \href {https://doi.org/10.1037/h0077714} {A circumplex model of affect}.
\newblock \emph{Journal of Personality and Social Psychology}, 39(6):1161--1178.

\bibitem[{Scherer and Wallbott(1994)}]{scherer1994isear}
Klaus~R. Scherer and Harald~G. Wallbott. 1994.
\newblock \href {https://www.unige.ch/cisa/research/materials-and-online-research/research-material/} {The {ISEAR} dataset: International survey on emotion antecedents and reactions}.
\newblock \emph{Technical Report}.

\bibitem[{Singh et~al.(2021)Singh, Joshi, Pawar, Chakrabarti, and Bhattacharyya}]{singh2021end}
Aaditya Singh, Aditya Joshi, Sachin Pawar, Soumen Chakrabarti, and Pushpak Bhattacharyya. 2021.
\newblock \href {https://doi.org/10.18653/v1/2021.eacl-main.125} {An end-to-end network for emotion-cause pair extraction}.
\newblock In \emph{Proceedings of the 16th Conference of the European Chapter of the Association for Computational Linguistics: Main Volume}, pages 1456--1461. Association for Computational Linguistics.

\bibitem[{Singh et~al.(2024)Singh, Caragea, and Li}]{singh-etal-2024-language}
Smriti Singh, Cornelia Caragea, and Junyi~Jessy Li. 2024.
\newblock \href {https://doi.org/10.18653/v1/2024.naacl-short.51} {Language models (mostly) do not consider emotion triggers when predicting emotion}.
\newblock In \emph{Proceedings of the 2024 Conference of the North American Chapter of the Association for Computational Linguistics: Human Language Technologies (Volume 2: Short Papers)}, pages 603--614, Mexico City, Mexico. Association for Computational Linguistics.

\bibitem[{Sosea and Caragea(2020)}]{sosea-caragea-2020-canceremo}
Tiberiu Sosea and Cornelia Caragea. 2020.
\newblock \href {https://doi.org/10.18653/v1/2020.emnlp-main.715} {{C}ancer{E}mo: A dataset for fine-grained emotion detection}.
\newblock In \emph{Proceedings of the 2020 Conference on Empirical Methods in Natural Language Processing (EMNLP)}, pages 8892--8904, Online. Association for Computational Linguistics.

\bibitem[{Strapparava and Mihalcea(2007)}]{strapparava-mihalcea-2007-semeval}
Carlo Strapparava and Rada Mihalcea. 2007.
\newblock \href {https://aclanthology.org/S07-1013} {{S}em{E}val-2007 task 14: Affective text}.
\newblock In \emph{Proceedings of the Fourth International Workshop on Semantic Evaluations ({S}em{E}val-2007)}, pages 70--74, Prague, Czech Republic. Association for Computational Linguistics.

\bibitem[{Team et~al.(2024)Team, Mesnard, Hardin, Dadashi, Bhupatiraju, Pathak, Sifre, Rivière, Kale, Love, Tafti, Hussenot, Sessa, Chowdhery, Roberts, Barua, Botev, Castro-Ros, Slone, Héliou, Tacchetti, Bulanova, Paterson, Tsai, Shahriari, Lan, Choquette-Choo, Crepy, Cer, Ippolito, Reid, Buchatskaya, Ni, Noland, Yan, Tucker, Muraru, Rozhdestvenskiy, Michalewski, Tenney, Grishchenko, Austin, Keeling, Labanowski, Lespiau, Stanway, Brennan, Chen, Ferret, Chiu, Mao-Jones, Lee, Yu, Millican, Sjoesund, Lee, Dixon, Reid, Mikuła, Wirth, Sharman, Chinaev, Thain, Bachem, Chang, Wahltinez, Bailey, Michel, Yotov, Chaabouni, Comanescu, Jana, Anil, McIlroy, Liu, Mullins, Smith, Borgeaud, Girgin, Douglas, Pandya, Shakeri, De, Klimenko, Hennigan, Feinberg, Stokowiec, hui Chen, Ahmed, Gong, Warkentin, Peran, Giang, Farabet, Vinyals, Dean, Kavukcuoglu, Hassabis, Ghahramani, Eck, Barral, Pereira, Collins, Joulin, Fiedel, Senter, Andreev, and Kenealy}]{gemma}
Gemma Team, Thomas Mesnard, Cassidy Hardin, Robert Dadashi, Surya Bhupatiraju, Shreya Pathak, Laurent Sifre, Morgane Rivière, Mihir~Sanjay Kale, Juliette Love, Pouya Tafti, Léonard Hussenot, Pier~Giuseppe Sessa, Aakanksha Chowdhery, Adam Roberts, Aditya Barua, Alex Botev, Alex Castro-Ros, Ambrose Slone, Amélie Héliou, Andrea Tacchetti, Anna Bulanova, Antonia Paterson, Beth Tsai, Bobak Shahriari, Charline~Le Lan, Christopher~A. Choquette-Choo, Clément Crepy, Daniel Cer, Daphne Ippolito, David Reid, Elena Buchatskaya, Eric Ni, Eric Noland, Geng Yan, George Tucker, George-Christian Muraru, Grigory Rozhdestvenskiy, Henryk Michalewski, Ian Tenney, Ivan Grishchenko, Jacob Austin, James Keeling, Jane Labanowski, Jean-Baptiste Lespiau, Jeff Stanway, Jenny Brennan, Jeremy Chen, Johan Ferret, Justin Chiu, Justin Mao-Jones, Katherine Lee, Kathy Yu, Katie Millican, Lars~Lowe Sjoesund, Lisa Lee, Lucas Dixon, Machel Reid, Maciej Mikuła, Mateo Wirth, Michael Sharman, Nikolai Chinaev, Nithum Thain, Olivier Bachem,
  Oscar Chang, Oscar Wahltinez, Paige Bailey, Paul Michel, Petko Yotov, Rahma Chaabouni, Ramona Comanescu, Reena Jana, Rohan Anil, Ross McIlroy, Ruibo Liu, Ryan Mullins, Samuel~L Smith, Sebastian Borgeaud, Sertan Girgin, Sholto Douglas, Shree Pandya, Siamak Shakeri, Soham De, Ted Klimenko, Tom Hennigan, Vlad Feinberg, Wojciech Stokowiec, Yu~hui Chen, Zafarali Ahmed, Zhitao Gong, Tris Warkentin, Ludovic Peran, Minh Giang, Clément Farabet, Oriol Vinyals, Jeff Dean, Koray Kavukcuoglu, Demis Hassabis, Zoubin Ghahramani, Douglas Eck, Joelle Barral, Fernando Pereira, Eli Collins, Armand Joulin, Noah Fiedel, Evan Senter, Alek Andreev, and Kathleen Kenealy. 2024.
\newblock \href {http://arxiv.org/abs/2403.08295} {Gemma: Open models based on gemini research and technology}.

\bibitem[{Wang et~al.(2022)Wang, Zhan, and Liu}]{wang2022}
Shanshan Wang, Lingling Zhan, and Rui Liu. 2022.
\newblock \href {https://doi.org/10.2139/ssrn.4054080} {Emotional responses in online reviews: The role of emotional clarity and intensity}.
\newblock \emph{SSRN Electronic Journal}.

\bibitem[{Wei et~al.(2020)Wei, Li, Zhang, Mao, He, and Wang}]{wei-etal-2020-unified}
Zhongyu Wei, Jiangxia Li, Yaojie Zhang, Zhendong Mao, Yulan He, and Bingquan Wang. 2020.
\newblock \href {https://doi.org/10.18653/v1/2020.coling-main.18} {A unified sequence labeling model for emotion cause pair extraction}.
\newblock In \emph{Proceedings of the 28th International Conference on Computational Linguistics}, pages 1449--1460, Barcelona, Spain (Online). International Committee on Computational Linguistics.

\bibitem[{Wolf et~al.(2020)Wolf, Debut, Sanh, Chaumond, Delangue, Moi, Cistac, Rault, Louf, Funtowicz, Davison, Shleifer, von Platen, Ma, Jernite, Plu, Xu, Le~Scao, Gugger, Drame, Lhoest, and Rush}]{hf}
Thomas Wolf, Lysandre Debut, Victor Sanh, Julien Chaumond, Clement Delangue, Anthony Moi, Pierric Cistac, Tim Rault, Remi Louf, Morgan Funtowicz, Joe Davison, Sam Shleifer, Patrick von Platen, Clara Ma, Yacine Jernite, Julien Plu, Canwen Xu, Teven Le~Scao, Sylvain Gugger, Mariama Drame, Quentin Lhoest, and Alexander Rush. 2020.
\newblock \href {https://doi.org/10.18653/v1/2020.emnlp-demos.6} {Transformers: State-of-the-art natural language processing}.
\newblock In \emph{Proceedings of the 2020 Conference on Empirical Methods in Natural Language Processing: System Demonstrations}, pages 38--45, Online. Association for Computational Linguistics.

\bibitem[{Xia and Ding(2019)}]{xia-ding-2019-emotion}
Rui Xia and Zixiang Ding. 2019.
\newblock \href {https://doi.org/10.18653/v1/P19-1096} {Emotion-cause pair extraction: A new task to emotion analysis in texts}.
\newblock In \emph{Proceedings of the 57th Annual Meeting of the Association for Computational Linguistics}, pages 1003--1012, Florence, Italy. Association for Computational Linguistics.

\bibitem[{Xie et~al.(2022)Xie, Jin, and Guo}]{Xie2022-rg}
Chunchang Xie, Junxi Jin, and Xiaoling Guo. 2022.
\newblock Impact of the critical factors of customer experience on well-being: Joy and customer satisfaction as mediators.
\newblock \emph{Front. Psychol.}, 13:955130.

\bibitem[{Yang et~al.(2024)Yang, Yang, Hui, Zheng, Yu, Zhou, Li, Li, Liu, Huang, Dong, Wei, Lin, Tang, Wang, Yang, Tu, Zhang, Ma, Yang, Xu, Zhou, Bai, He, Lin, Dang, Lu, Chen, Yang, Li, Xue, Ni, Zhang, Wang, Peng, Men, Gao, Lin, Wang, Bai, Tan, Zhu, Li, Liu, Ge, Deng, Zhou, Ren, Zhang, Wei, Ren, Liu, Fan, Yao, Zhang, Wan, Chu, Liu, Cui, Zhang, Guo, and Fan}]{q}
An~Yang, Baosong Yang, Binyuan Hui, Bo~Zheng, Bowen Yu, Chang Zhou, Chengpeng Li, Chengyuan Li, Dayiheng Liu, Fei Huang, Guanting Dong, Haoran Wei, Huan Lin, Jialong Tang, Jialin Wang, Jian Yang, Jianhong Tu, Jianwei Zhang, Jianxin Ma, Jianxin Yang, Jin Xu, Jingren Zhou, Jinze Bai, Jinzheng He, Junyang Lin, Kai Dang, Keming Lu, Keqin Chen, Kexin Yang, Mei Li, Mingfeng Xue, Na~Ni, Pei Zhang, Peng Wang, Ru~Peng, Rui Men, Ruize Gao, Runji Lin, Shijie Wang, Shuai Bai, Sinan Tan, Tianhang Zhu, Tianhao Li, Tianyu Liu, Wenbin Ge, Xiaodong Deng, Xiaohuan Zhou, Xingzhang Ren, Xinyu Zhang, Xipin Wei, Xuancheng Ren, Xuejing Liu, Yang Fan, Yang Yao, Yichang Zhang, Yu~Wan, Yunfei Chu, Yuqiong Liu, Zeyu Cui, Zhenru Zhang, Zhifang Guo, and Zhihao Fan. 2024.
\newblock \href {http://arxiv.org/abs/2407.10671} {Qwen2 technical report}.

\bibitem[{Yelp(2025)}]{yelp_dataset}
Yelp. 2025.
\newblock \href {https://business.yelp.com/data/resources/open-dataset/} {Yelp open dataset}.
\newblock Accessed: 2025-02-16.

\bibitem[{Yin et~al.(2014)Yin, Bond, and Zhang}]{abc}
Dezhi Yin, Samuel~D. Bond, and Han Zhang. 2014.
\newblock \href {https://doi.org/10.25300/MISQ/2014/38.2.10} {Anxious or angry? effects of discrete emotions on the perceived helpfulness of online reviews}.
\newblock \emph{MIS Q.}, 38(2):539–560.

\bibitem[{Yuan et~al.(2020)Yuan, Zhang, and Liu}]{yuan2020ecpe}
Yujie Yuan, Min Zhang, and Yang Liu. 2020.
\newblock \href {https://doi.org/10.18653/v1/2020.acl-main.91} {Emotion-cause pair extraction with graph convolutional networks}.
\newblock In \emph{Proceedings of the 58th Annual Meeting of the Association for Computational Linguistics}, pages 1010--1020, Online. Association for Computational Linguistics.

\end{thebibliography}

\appendix
\section{LLMs Utilized}\label{llms_list}
In our experiments, we adopt a range of recent widely-used LLMs. For close-sourced LLM (accessible through APIs), we evaluate OpenAI’s models   \cite{openai}. For open-source LLMs, use the HuggingFace library \cite{hf} to access these models and experimented with Llama family of models \cite{l3}, {\gemmaone} \cite{gemma}, {\gemmatwo} \cite{gemma}, {\mistralinstructv{7}{0.2}} \citet{m2}, {\mistralinstructv{7}{0.3}} \cite{m2}, {\qweninstructnew} \cite{q} and {\mixtralnew} \cite{mix}.

\section{Inference Configuration Details}

\begin{table}[t]
    \centering
    \resizebox{\columnwidth}{!}{%
    
    \begin{tabular}{lcccccc}
    \toprule
         \textbf{Metric} & \textbf{Beauty} & \textbf{Clothing} & \textbf{Home} & \textbf{Electronics} & \textbf{TripAdvisor} & \textbf{Yelp} \\
    \midrule
        \multicolumn{7}{c}{\textit{Overall Statistics}}\\
    \midrule
        
        Total Emotions & 957 & 897 & 916 & 913 & 1183 & 1084 \\
        Avg Emotions per Review & 2.39 & 2.24 & 2.29 & 2.28 & 2.96 & 2.71 \\
    \midrule
        \multicolumn{7}{c}{\textit{Emotion: Anger}}\\
    \midrule
        Count (Percentage) & 58 (6.1\%) & 41 (4.6\%) & 73 (8.0\%) & 82 (9.0\%) & 86 (7.3\%) & 66 (6.1\%) \\
        Total Triggers & 148 & 106 & 221 & 261 & 371 & 270 \\
        Avg Triggers per Emotion & 2.55 & 2.59 & 3.03 & 3.18 & 4.31 & 4.09 \\
        Avg Trigger Length (words) & 6.80 & 8.70 & 8.20 & 8.52 & 9.73 & 9.12 \\
    \midrule
        \multicolumn{7}{c}{\textit{Emotion: Anticipation}}\\
    \midrule
        Count (Percentage) & 105 (11.0\%) & 100 (11.1\%) & 82 (9.0\%) & 89 (9.7\%) & 180 (15.2\%) & 196 (18.1\%) \\
        Total Triggers & 148 & 135 & 122 & 133 & 279 & 300 \\
        Avg Triggers per Emotion & 1.41 & 1.35 & 1.49 & 1.49 & 1.55 & 1.53 \\
        Avg Trigger Length (words) & 10.30 & 10.62 & 10.24 & 11.23 & 10.62 & 10.40 \\
    \midrule
        \multicolumn{7}{c}{\textit{Emotion: Disgust}}\\
    \midrule
        Count (Percentage) & 139 (14.5\%) & 111 (12.4\%) & 117 (12.8\%) & 120 (13.1\%) & 172 (14.5\%) & 101 (9.3\%) \\
        Total Triggers & 292 & 215 & 256 & 277 & 506 & 268 \\
        Avg Triggers per Emotion & 2.10 & 1.94 & 2.19 & 2.31 & 2.94 & 2.65 \\
        Avg Trigger Length (words) & 7.81 & 8.65 & 8.38 & 9.02 & 8.92 & 9.10 \\
    \midrule
        \multicolumn{7}{c}{\textit{Emotion: Fear}}\\
    \midrule
        Count (Percentage) & 14 (1.5\%) & 4 (0.4\%) & 16 (1.7\%) & 16 (1.8\%) & 36 (3.0\%) & 12 (1.1\%) \\
        Total Triggers & 18 & 5 & 20 & 28 & 64 & 20 \\
        Avg Triggers per Emotion & 1.29 & 1.25 & 1.25 & 1.75 & 1.78 & 1.67 \\
        Avg Trigger Length (words) & 13.00 & 12.80 & 10.30 & 9.71 & 11.12 & 10.80 \\
    \midrule
        \multicolumn{7}{c}{\textit{Emotion: Joy}}\\
    \midrule
        Count (Percentage) & 294 (30.7\%) & 318 (35.5\%) & 305 (33.3\%) & 282 (30.9\%) & 312 (26.4\%) & 328 (30.3\%) \\
        Total Triggers & 829 & 846 & 809 & 716 & 1498 & 1353 \\
        Avg Triggers per Emotion & 2.82 & 2.66 & 2.65 & 2.54 & 4.80 & 4.12 \\
        Avg Trigger Length (words) & 7.04 & 6.43 & 6.97 & 7.14 & 8.14 & 7.86 \\
    \midrule
        \multicolumn{7}{c}{\textit{Emotion: Sadness}}\\
    \midrule
        Count (Percentage) & 112 (11.7\%) & 100 (11.1\%) & 103 (11.2\%) & 103 (11.3\%) & 95 (8.0\%) & 76 (7.0\%) \\
        Total Triggers & 221 & 204 & 234 & 201 & 262 & 167 \\
        Avg Triggers per Emotion & 1.97 & 2.04 & 2.27 & 1.95 & 2.76 & 2.20 \\
        Avg Trigger Length (words) & 7.91 & 8.80 & 7.84 & 8.79 & 9.57 & 10.16 \\
    \bottomrule
    \end{tabular}%
    }
    \caption{Gold Standard Aggregated Dataset: Emotion Statistics across Domains.}
    \label{tab:gold_standard}
\end{table}

For our experiments, we implemented a standardized inference configuration across both closed-source and open-source LLMs to ensure consistent and reliable outputs. While different LLM providers recommend varying default settings (e.g., {\texttt GPT-4o} suggests \texttt{temperature=1.0}, {\texttt DeepSeek} recommends (\texttt{temperature=0.6}), we conducted extensive parameter tuning to optimize for our specific task requirements.

\subsection{Parameter Settings}
We empirically determined the following optimal configuration:
\texttt{inference\_config = \{'temperature': 0.2, 'top\_p': 0.95, 'top\_k': 25, 'max\_tokens': 2500, 'n': 1\}}

\subsection{Rationale for Parameter Choices}

\textls[-60]{\textbf{\tt Temperature} = $0.2$}
: We deliberately chose a low temperature setting for several reasons:
\begin{itemize}
    \item Our task focuses on \textit{detection} and \textit{extraction} rather than creative generation.
    \item Lower temperature produces more deterministic outputs, which is crucial for consistent emotion detection.
    \item Reduces variation in \textit{trigger span} identification, leading to more reliable extractions.
    \item Aligns with our need for \textit{precise}, focused responses rather than diverse creative generations.
\end{itemize}

\textbf{{\textsc{Top-p} (0.95) and \textsc{Top-k} (25)}}: These settings provide a balanced approach to sampling:
\begin{itemize}
    \item \texttt{top-p = 0.95} ensures consideration of the most probable tokens while maintaining some flexibility.
    \item \texttt{top-k = 25} limits the selection pool to the most relevant tokens.
    \item The combination helps maintain coherence while capturing \textit{emotional nuances} in the reviews.
\end{itemize}

{\textsc{\textbf{Max Tokens}} ($\textbf{2500}$)}: This limit was set to accommodate:
\begin{itemize}
    \item Comprehensive \textit{emotion analysis}.
    \item Multiple \textit{trigger span} extractions.
    \item \textit{Reasoning steps} in the chain-of-thought prompting.
    \item \textit{Self-reflection} mechanisms in the \textsc{EOT-DETECT} framework.
\end{itemize}

These parameters were validated through extensive testing across our evaluation suite, demonstrating consistent performance in both \textit{emotion detection} and \textit{trigger extraction} tasks. The configuration prioritizes \textit{reliability} and \textit{precision} over creativity, aligning with our task's objective of accurate analysis rather than generative capability.

\section{Implementation Details of Fine-tuned Models}

We provide comprehensive details about our experimental setup, including data preprocessing, model architecture, and training configuration. All experiments were conducted using the Unsloth framework.

\subsection{Data Preprocessing}
We randomly split EOT-X into training (80\%), validation (10\%), and test (10\%) sets, ensuring that reviews from the same product (for Amazon) or experience/service (for TripAdvisor and Yelp) were kept within the same split to prevent data leakage.

\begin{table}[H]
    \centering
    \resizebox{1.0\columnwidth}{!}{%
    \begin{tabular}{l l}
    \toprule
         \textbf{Hyperparameter}        & \textbf{Value}                 \\
    \midrule
        \multicolumn{2}{l}{\textit{Dataset Statistics}} \\
        \texttt{Training Set}    & 1,919 examples                    \\
        \texttt{Validation Set}  & 239 examples                      \\
        \texttt{Test Set}        & 241 examples                      \\
    \midrule
        \multicolumn{2}{l}{\textit{Model Architecture}} \\
        \texttt{Maximum Sequence Length} & 2,048 tokens                   \\
        \texttt{Data Type} & bfloat16                                \\
    \midrule
        \multicolumn{2}{l}{\textit{LoRA Configuration}} \\
        \texttt{Rank (r)}    & 128                                   \\
        \texttt{Alpha}       & 128                                   \\
        \texttt{Dropout}     & 0.05                                  \\
        \texttt{Use RSLoRA}  & True                                  \\
    \midrule
        \multicolumn{2}{l}{\textit{Training Configuration}} \\
        \texttt{Number of Epochs}  & 5                              \\
        \texttt{Batch Size (per device)} & 2                         \\
        \texttt{Gradient Accumulation Steps} & 8                     \\
        \texttt{Effective Batch Size} & 16                            \\
        \texttt{Total Training Steps} & 600                           \\
        \texttt{Optimizer} & AdamW                                    \\
        \texttt{Learning Rate} & 2e-5                                 \\
        \texttt{Weight Decay} & 0.01                                  \\
        \texttt{Learning Rate Scheduler} & Cosine                    \\
        \texttt{Warmup Ratio} & 0.05                                  \\
        \texttt{Early Stopping Patience} & 2                          \\
        \texttt{Early Stopping Threshold} & 0.005                    \\
        \texttt{Save/Evaluation Frequency} & Every 200 steps          \\
    \midrule
        \multicolumn{2}{l}{\textit{Inference Configuration}} \\
        \texttt{Maximum New Tokens} & 2,048                          \\
        \texttt{Temperature} & 0.2                                   \\
        \texttt{Top-p} & 0.95                                        \\
        \texttt{Top-k} & 25                                          \\
        \texttt{Repetition Penalty} & 1.1                            \\
    \bottomrule
    \end{tabular}
    }
    \caption{Hyperparameters and configuration details for our experiments.}
    \label{tab:hyperparameters}
\end{table}

\section{Annotation Guidelines}\label{guidelines}

{\textbf{Introduction for Annotators}}

\textit{Dear Annotators,}

Thank you for participating in this crucial annotation task for \textsc{Emotion and Opinion Trigger Detection} in e-commerce reviews. Your expertise is vital for constructing the high-quality \textsc{EOT-X} dataset, which will significantly advance our understanding of customer emotional responses and their underlying triggers in product reviews.

In this task, you will identify emotions expressed in reviews and extract the exact text spans (\textsc{opinion triggers}) that explain why each emotion is present. Please read these guidelines thoroughly before beginning the annotation process.

\subsection{Task Overview}
Your annotation task consists of two main components:

\noindent\textbf{Emotion Detection:} Identify all emotions expressed in a review using \textsc{Plutchik's} 8 primary emotions.\\
\textbf{Opinion Trigger Extraction:} Extract the precise text spans from the review that explain the presence of each detected emotion.

\subsection{Emotion Framework: \textsc{Plutchik's} 8 Primary Emotions}
Focus exclusively on these 8 primary emotions:

\begin{itemize}
    \item \textls[-60]{Joy}
    \item \textls[-60]{Trust}
    \item \textls[-60]{Fear}
    \item \textls[-60]{Surprise}
    \item \textls[-60]{Sadness}
    \item \textls[-60]{Disgust}
    \item \textls[-60]{Anger}
    \item \textls[-60]{Anticipation}
\end{itemize}

\textbf{Note:} If no clear emotion is detected in the review, label it as \textit{Neutral}.

\subsection{Opinion Triggers}
An \textsc{opinion trigger} is defined as a direct extract from the review that explains why the customer experienced a particular emotion. Each trigger must be:

\begin{itemize}
    \item \textbf{Extractive:} Exactly as it appears in the review.
    \item \textbf{Clearly Linked:} Directly associated with the identified emotion.
    \item \textbf{Self-Contained:} Understandable without needing additional context.
\end{itemize}

\subsection{Annotation Process}
\textbf{General Instructions}
\begin{itemize}
    \item \textbf{Read Thoroughly:} Carefully read the entire review before starting your annotations.
    \item \textbf{Identify Emotions:} Detect all emotions expressed in the review.
    \item \textbf{Extract Triggers:} For each detected emotion, extract all relevant opinion triggers as exact text spans.
    \item \textbf{Documentation:} Record your annotations in the provided format.
    \item \textbf{Confidence Level:} Only mark emotions and triggers you are confident about.
\end{itemize}

\subsection{Specific Guidelines for Emotion Detection}
\begin{itemize}
    \item \textbf{Multiple Emotions:} Reviews may express more than one emotion. Annotate every clearly expressed emotion.
    \item \textbf{Emotion Intensity:} Focus solely on the presence or absence of an emotion; do not attempt to annotate the intensity.
    \item \textbf{Implicit vs. Explicit:} Annotate only those emotions that are clearly expressed. Do not infer or speculate.
    \item \textbf{Contextual Consideration:} Use the entire review context to guide your decision.
\end{itemize}

\paragraph{Edge Cases and Special Considerations}
Trigger Boundary Cases:

\textbf{Nested Triggers:} When a shorter trigger is nested within a longer one, choose the span that is most informative and coherent.\\
\textbf{Discontinuous Triggers:} Do not annotate discontinuous spans. Choose the most representative continuous span that best explains the emotion.\\
\textbf{Comparative Expressions:} For comparative statements (e.g., \textit{Better than all other brands I've tried}), include the full comparison if it is directly relevant to explaining the emotion.

\subsection{Special Review Types}
\textbf{Very Short Reviews:} Even in very short reviews, if clear emotional signals are present, annotate them. In such cases, the entire review might serve as the trigger.\\
\textbf{Technical Reviews:} Focus on the emotional content, not on technical details, unless the technical detail directly causes an emotional reaction.\\
\textbf{Sarcastic Reviews:} Identify and annotate the intended emotion rather than the literal meaning. If sarcasm is detected, note it in the comments field (if needed).

\subsection{Annotation Format}
For each review, provide the following information:

\begin{itemize}
    \item \textbf{Review ID}
    \item \textbf{List of Detected Emotions:} For each emotion, include:
    \begin{itemize}
        \item \textbf{Emotion Label:} One of \textsc{Plutchik's} 8 primary emotions or \textit{Neutral}.
        \item \textbf{List of Triggers:} Exact text spans from the review.
    \end{itemize}
\end{itemize}

\subsection{Format Example}
\begin{verbatim}
{
  "review_id": "123",
  "emotions": [
    {
      "emotion": "Joy",
      "triggers": [
        "works perfectly every time",
        "excellent battery life"
      ]
    }
  ]
}
\end{verbatim}

\paragraph{Quality Control Measures}
Self-Check Before Submission:
\begin{itemize}
    \item \textbf{Emotion Consistency:} All annotated emotions belong to \textsc{Plutchik's} set or are labeled as Neutral.
    \item \textbf{Trigger Accuracy:} Every \textsc{opinion trigger} is an exact text span from the review.
    \item \textbf{Clear Linkage:} Each emotion-trigger pair is clearly and unambiguously connected.
    \item \textbf{Completeness:} No clear emotional expressions are left unannotated.
\end{itemize}

\paragraph{Final Notes}
(\textit{1}) \textbf{Attention to Detail:} Your meticulous attention to detail is crucial for the success of this dataset.
(\textit{2}) \textbf{Feedback:} If you encounter any ambiguities or issues, please reach out to the research team immediately.
(\textit{3}) \textbf{Consistency:} Maintain high consistency across all annotations to ensure the dataset is robust and reliable.
(\textit{4}) \textbf{Expert Judgment:} Rely on your expertise in both emotion analysis and consumer psychology to make nuanced decisions.

\textit{Thank you for your dedication and careful work in this annotation task. Your contributions are instrumental in advancing research on emotion-opinion trigger detection in e-commerce reviews.}

\section{Model Responses}

\begin{table}[t] 
\centering
\caption{Deepseek-R1 Model Response}
\label{tab:response1}
\begin{tabular}{p{0.95\columnwidth}}  
\toprule
\textbf{Model:} Deepseek-R1 \\
[0.2cm]
\textbf{Response:} \\
[0.2cm]
\textbf{Emotions:} \\
\textbf{Trust} \\
Triggers: \\
\quad • The only shoes she will wear to work \\
\quad • Non slip, great arch support \\
\midrule
\textbf{Anticipation} \\
Triggers: \\
\quad • She can wear thicker socks in winter \\
\quad • In summer she has room in case her feet swell a little \\
\bottomrule
\end{tabular}
\end{table}

\begin{table}[t] 
\centering
\caption{GPT-4o Model Response}
\label{tab:response1}
\begin{tabular}{p{0.95\columnwidth}}  
\toprule
\textbf{Model:} GPT-4o \\
[0.2cm]
\textbf{Response:} \\
[0.2cm]
\textbf{Emotions:} \\
\textbf{Joy} \\
Triggers: \\
\quad • I've been buying these for my mom for 2 years now \\
\quad • These are the only shoes she will wear to work \\
\quad • Non slip, great arch support \\
\midrule
\textbf{Trust} \\
Triggers: \\
\quad • These are the only shoes she will wear to work \\
\quad • Non slip, great arch support \\
\quad • A touch wider than normal shoes (like most Skechers) so she can wear thicker socks in winter and in summer she has room in case her feet swell a little \\
\midrule
\textbf{Anticipation} \\
Triggers: \\
\quad • My mom replaces about every 6 months as she works 6 days a week \\
\bottomrule
\end{tabular}
\end{table}

\begin{table}[t] 
\centering
\caption{o1-mini Model Response }
\label{tab:response1}
\begin{tabular}{p{0.95\columnwidth}}  
\toprule
\textbf{Model:} o1-mini \\
[0.2cm]
\textbf{Response:} \\
[0.2cm]
\textbf{Emotions:} \\
\textbf{Trust} \\
Triggers: \\
\quad • These are the only shoes she will wear to work \\
\quad • Non slip \\
\quad • Great arch support \\
\quad • She replaces about every 6 months as she works 6 days a week \\
\quad • A touch wider than normal shoes \\
\quad • She can wear thicker socks in winter and in summer she has room in case her feet swell a little \\
\midrule
\textbf{Joy} \\
Triggers: \\
\quad • I've been buying these for my mom for 2 years now \\
\quad • These are the only shoes she will wear to work \\
\midrule
\textbf{Anticipation} \\
Triggers: \\
\quad • She replaces about every 6 months as she works 6 days a week \\
\bottomrule
\end{tabular}
\end{table}

\begin{table}[t] 
\centering
\caption{Llama3.3 70B-INSTRUCT Model Response}
\label{tab:response1}
\begin{tabular}{p{0.95\columnwidth}}  
\toprule
\textbf{Model:} Llama3.3 70B \\
[0.2cm]
\textbf{Response:} \\
[0.2cm]
\textbf{Emotions:} \\
\textbf{Trust} \\
Triggers: \\
\quad • The only shoes she will wear to work \\
\quad • Great arch support \\
\midrule
\textbf{Joy} \\
Triggers: \\
\quad • She can wear thicker socks in winter and in summer she has room in case her feet swell a little \\
\midrule
\textbf{Anticipation} \\
Triggers: \\
\quad • My mom replaces about every 6 months as she works 6 days a week \\
\bottomrule
\end{tabular}
\end{table}

\begin{table}[t] 
\centering
\caption{Mistral 2501 Model Response}
\label{tab:response1}
\begin{tabular}{p{0.95\columnwidth}}  
\toprule
\textbf{Model:} Mistral 2501 \\
[0.2cm]
\textbf{Response:} \\
[0.2cm]
\textbf{Emotions:} \\
\textbf{Joy} \\
Triggers: \\
\quad • She works in food service, on her feet 10+ hours at a time and these are the only shoes she will wear to work! \\
\quad • Non slip, great arch support; my mom replaces about every 6 months as she works 6 days a week. \\
\midrule
\textbf{Trust} \\
Triggers: \\
\quad • I've been buying these for my mom for 2 years now. \\
\midrule
\textbf{Anticipation} \\
Triggers: \\
\quad • A touch wider than normal shoes (like most Skechers) so she can wear thicker socks in winter and in summer she has room in case her feet swell a little. \\
\bottomrule
\end{tabular}
\end{table}

\begin{table}[t] 
\centering
\caption{Claude 3.5 Sonnet Model Response}
\label{tab:response1}
\begin{tabular}{p{0.95\columnwidth}}  
\toprule
\textbf{Model:} Claude 3.5 Sonnet \\
[0.2cm]
\textbf{Response:} \\
[0.2cm]
\textbf{Emotions:} \\
\textbf{Trust} \\
Triggers: \\
\quad • I've been buying these for my mom for 2 years now \\
\quad • These are the only shoes she will wear to work \\
\quad • My mom replaces about every 6 months \\
\midrule
\textbf{Joy} \\
Triggers: \\
\quad • Non slip, great arch support \\
\quad • She has room in case her feet swell a little \\
\bottomrule
\end{tabular}
\end{table}

\end{document}